%% file: root.tex
\documentclass[letterpaper, 10 pt, conference]{Style/ieeeconf}  

\IEEEoverridecommandlockouts                              

\input{packages}
\input{macros}

\title{Effect of Gait Design on Proprioceptive Sensing of Terrain Properties in a Quadrupedal Robot  

}

\setboolean{BLIND_REVIEW_FLAG}{false}
\ifthenelse{\boolean{BLIND_REVIEW_FLAG}}{
    \author{Author Names Omitted for Anonymous Review}
    }{
    \author{\efName$^{\star\dagger}$, \jdcName$^{\diamond\dagger}$, \yzName$^{\diamond}$, \jrName$^{\circ}$, 
    and \fqName$^{\diamond}$%
    \thanks{$^{\star}$\mitME. 
    {\tt\small \efUSCEmail@mit.edu}}
    \thanks{$^{\diamond}$\uscECE. 
    {\tt\small \{\jdcUSCEmail,\yzUSCEmail,\fqUSCEmail\}@usc.edu}}
    \thanks{$^{\circ}$\sasDep. 
    {\tt\small \jrUPennEmail}}
    \thanks{$^{\dagger}$ Equal Contribution}
    }}

\begin{document}
\listoftodos
\maketitle
\bstctlcite{IEEEexample:BSTcontrol}
\setcounter{page}{1}

\begin{abstract}
In-situ robotic exploration is an important tool for advancing knowledge of geological processes that describe the Earth and other Planetary bodies. 
To inform and enhance operations for these roving laboratories, it is imperative to understand the terramechanical properties of their environments, especially for traversing on loose, deformable substrates.
Recent research suggested that legged robots with direct-drive and low-gear ratio actuators can sensitively detect external forces, and therefore possess the potential to measure terrain properties with their legs during locomotion, providing unprecedented sampling speed and density while accessing terrains previously too risky to sample. 
This paper explores these ideas by investigating the impact of gait on proprioceptive terrain sensing accuracy, particularly comparing a sensing-oriented gait, \crawlsense, with a locomotion-oriented gait, \trotloco. 
Each gait's ability to measure the strength and texture of deformable substrate is quantified as the robot locomotes over a laboratory transect consisting of a rigid surface, loose sand, and loose sand with synthetic surface crusts. 
Our results suggest that with both the sensing-oriented crawling gait and locomotion-oriented trot gait, the robot can measure a consistent difference in the strength (in terms of penetration resistance) between the low- and high-resistance substrates; however, the locomotion-oriented trot gait contains larger magnitude and variance in measurements. 
Furthermore, the slower crawl gait can detect brittle ruptures of the surface crusts with significantly higher accuracy than the faster trot gait. 
Our results offer new insights that inform legged robot ``sensing during locomotion'' gait design and planning for scouting the terrain and producing scientific measurements on other worlds to advance our understanding of their geology and formation. 

\end{abstract}

\input{Sections/Introduction}
\input{Sections/FieldResults}
\input{Sections/Methods}
\input{Sections/Results}
\input{Sections/ConclusionDiscussions}


\section*{Acknowledgment}
This work is supported by the NASA Mars Exploration Program (MEP) Technology Development Funding, the NASA Planetary Science and Technology Through Analog Research (PSTAR) program, Award \#80NSSC22K1313, the NASA Lunar Surface Technology Research (LuSTR) program, Award \#80NSSC24K0127, and the National Science Foundation (NSF) CAREER award \#2240075.

\bibliographystyle{Style/IEEEtran}
\bibliography{bst_control, Style/IEEEabrv, allRefs}

\end{document}

%% file: packages.tex
\usepackage[nospace]{cite} 
\usepackage{amsmath,amssymb,amsfonts}
\usepackage{algorithmic}
\usepackage{graphicx}
\usepackage{textcomp}
\usepackage[table]{xcolor} 
\usepackage{hyperref}
\usepackage{cancel}
\usepackage{ulem}
\usepackage[capitalize]{cleveref}
\usepackage{subcaption}
\usepackage{transparent} 
\usepackage{etoolbox} 

\DeclareCaptionLabelFormat{periodAfter}{#2.}
\crefname{section}{Sec.}{Secs.}

\usepackage{mathtools}
\usepackage{overpic}
\usepackage{ifthen}
\usepackage{xspace}
\usepackage{array}
\usepackage{tabularx}

\newcolumntype{Y}{>{\centering\arraybackslash}X}
\newcolumntype{Z}[1]{>{\centering\arraybackslash}m{#1}}
\newcolumntype{?}{!{\vrule width 1.5pt}}

\hypersetup{
    colorlinks,
    linkcolor={red},
    citecolor={blue},
    urlcolor={blue}
}

\usepackage[disable]{todonotes}
\usepackage{xparse}

%% file: macros.tex
\newcommand{\jdcName}{J. Diego Caporale}
\newcommand{\jdcUSCEmail}{caporale}

\newcommand{\efName}{Ethan Fulcher}

\newcommand{\yzName}{Yifeng Zhang}
\newcommand{\yzUSCEmail}{yifengz}

\newcommand{\fqName}{Feifei Qian}
\newcommand{\fqUSCEmail}{feifeiqi}

\newcommand{\jrName}{John Ruck}
\newcommand{\jrUPennEmail}{johnruck@sas.upenn.edu}

\newcommand{\uscECE}{Ming Hsieh Department of Electrical and Computer Engineering, University of Southern California, Los Angeles, CA, USA}

\newcommand{\mitME}{Department of Mechanical Engineering, Massachusetts Institute of Technology, Los Angeles, CA, USA}
\newcommand{\sasDep}{Department of Earth and Environmental Science, University of Pennsylvania, Philadelphia, PA, USA}

\makeatletter
\NewDocumentCommand{\todotext}{O{red} O{} m}{%
  \@ifpackagewith{todonotes}{disable}
    {}
    {%
      { \xspace \textcolor{#1}{#2} \textcolor{#1}{#3} \xspace}%
      \todo[
          noline,
          color=#1,
          bordercolor=black,
          linecolor=white,
          textcolor=white,
          caption={#2 #3}
        ]{{.}}%
      }%
    }%
\makeatother

\def\eg{\textit{e.g.},\ }
\def\ie{ \textit{i.e.},\ }

\def\vs{\textit{v.s.\/ }}
\def\insitu{\textit{in\,situ}\xspace}

\def\stiffness{k}

\DeclareRobustCommand{\crawlsense}{\textit{Crawl~N'~Sense}\xspace} 
\DeclareRobustCommand{\trotloco}{\textit{Trot\mbox{-}Walk}\xspace} 

\def\frameGround{G}
\def\normalGround{\hat{z}_\frameGround}

\def\timeROI{\mathbf{t}_{roi}}

\newboolean{BLIND_REVIEW_FLAG} 

%% file: Sections/Introduction.tex
\section{Introduction}
Robotic exploration plays a crucial role in advancing our understanding of Earth and other planetary bodies by enabling \insitu experiments remotely\cite{https://doi.org/10.1029/2002JE002038}. 
Many terrestrial and planetary environments (\cref{fig:intro_travleg,fig:intro_wheel,fig:fieldOver}) present significant challenges for traditional wheeled rovers\cite{doi:10.2514/6.2005-2525}, as hazardous terrains limit access to areas of scientific interest. 
Given the high mission costs and risks, scientists must be risk-averse, often foregoing scientific opportunities that pose potential threats to robotic platforms. 
This is partly due to the challenge in acquiring terramechanical information about surface properties such as regolith strength and texture\cite{CHHANIYARA2012115}, which are difficult to infer without direct tactile feedback. 
Traditional methods\cite{CHHANIYARA2012115} require dedicated sensors that demand stopping the rover for measurements, leading to sparse data collection and operational inefficiencies.

\input{Figures/1-Motivation/motivation.fig}

Legged robots\cite{saranli2001rhex,kenneally2016design,katz2019mini,boston_dynamics_spot,arm2019SpaceBokDynamicLegged} offer an alternative mobility paradigm that can expand the operational envelope of planetary exploration. 
Recent work has demonstrated their potential not only for traversing extreme terrain\cite{ilhan2018autonomous,qian2019rapid,arm2023scientificexploration,hoeller2024anymalparkour} but also for proprioceptively ``feeling" surface interactions as they walk, effectively acting as penetrometers and making every step an experiment. 
Recent advances in actuator transparency and torque density have led to accurate force estimation in direct-drive (\ie gearless)~\cite{kenneally2018actuator} and quasi-direct-drive (QDD, \ie low gear ratio) robotic limbs (\cref{fig:intro_travleg,fig:intro_spirit}), which has shown promise in enabling high-density ground reaction force measurements~\cite{qian2019rapid,ruck2023downslope}. 
However, most prior work has focused on static or slow-moving test setups\cite{qian2017ground,qian2019rapid,bush2023robotic,ruck2023downslope,liu2025adaptive} (\cref{fig:intro_travleg,fig:intro_gaits}.1). 
Here, we explore the possibility for proprioceptive terrain sensing to be effectively conducted while a legged robot is in motion (\cref{fig:intro_spirit,fig:intro_gaits}.2-3). 
This capability would allow every step to serve as a scientific experiment, significantly increasing the density and speed of geotechnical surveys and planetary explorations. 
From a scientific perspective, this could reveal spatial gradients in surface properties\cite{jerolmack2006spatial,jerolmack2011sorting,kostynick2022rheology}, providing deeper insight into planetary surface processes\cite{qian2019rapid}. 
Operationally, it could enhance mission planning by informing rover and astronaut activities such as excavation, sampling, navigation\cite{Keenan2024CADRE}, and construction \cite{holladay2025LunarConstruction}.

To achieve this, a key question that needs to be answered is: how does gait design affect the accuracy and coverage of proprioceptive regolith sensing? 
High-speed locomotion enables rapid data collection but introduces complex regolith reaction force profiles~\cite{qian2013walking} that can complicate signal interpretation, potentially reducing measurement accuracy. 
Conversely, low-speed gaits, which keep footsteps in the quasi-static regime, provide higher confidence in force estimation but limit spatial coverage. 

To answer these questions, this study examines how variations in gait design influence a QDD quadrupedal robot's ability to measure terrain properties such as strength and texture change.
Field experiments with a mounted leg penetrometer (\cref{fig:intro_travleg,fig:intro_gaits}.1), in an analogue environment (\cref{fig:MH}) motivate the need for sensitive, dense measurement of the terrain strength (\cref{sec:fieldresults}).
Furthermore, we develop a custom sensing-oriented gait, the  \crawlsense (\cref{fig:intro_gaits}.2), and compare it to a baseline locomotion-oriented gait, the \trotloco (\cref{fig:intro_gaits}.3), in 
lab experiments (\cref{sec:methods}). 
As the robot traverses varied terrains, we characterize the proprioceptive joint signals and propose methods to extract terrain properties from these signals, including a method for proprioceptively estimating the ground surface on soft terrain via a contact-based correction (\cref{sec:EstimatingDepth}).
Using the proposed methods, we systematically compare the two gaits in terms of sensing accuracy for (i) characterizing regolith strength (\cref{sec:stiffness-detection}) and (ii) detecting texture and layering (\cref{sec:crust-detection}).
Our results provide key insights into the gait design principles needed to optimize both mobility and sensing in legged robots, with implications for deployment in complex earth and planetary surface explorations. 

%% file: Figures/1-Motivation/motivation.fig.tex
\begin{figure}[t]
\centering 
\vspace{6pt}
\begin{subfigure}[t]{.99\linewidth}
\centering
    \begin{overpic}[height=0.35\linewidth]{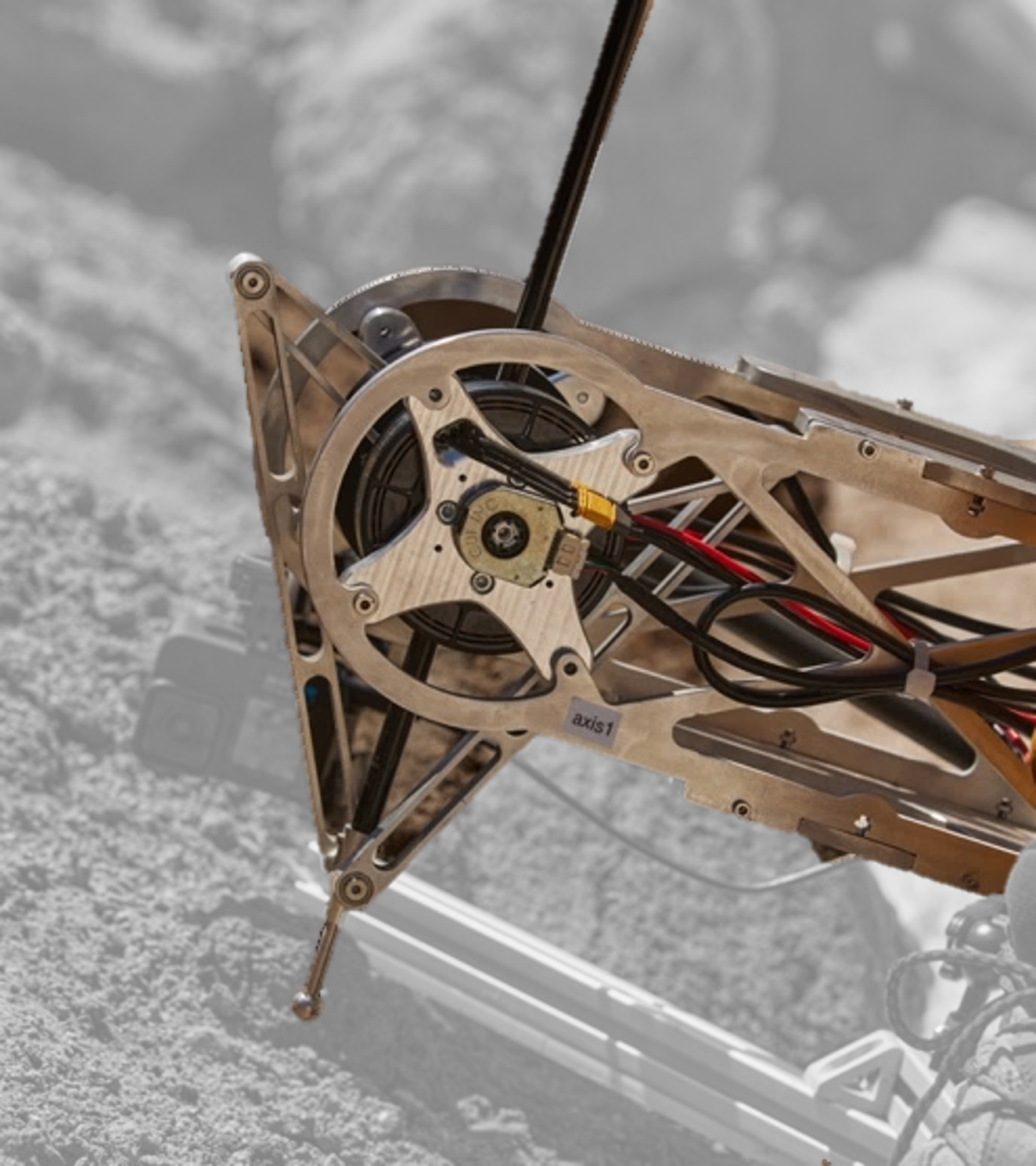}%
        {%
        \setlength{\fboxsep}{2pt}%
        \put(2, 5){\colorbox{white}{\textbf{\footnotesize{A}}}}
        }%
    \end{overpic}
        \phantomcaption
        \label{fig:intro_travleg}
    \begin{overpic}[height=0.35\linewidth]{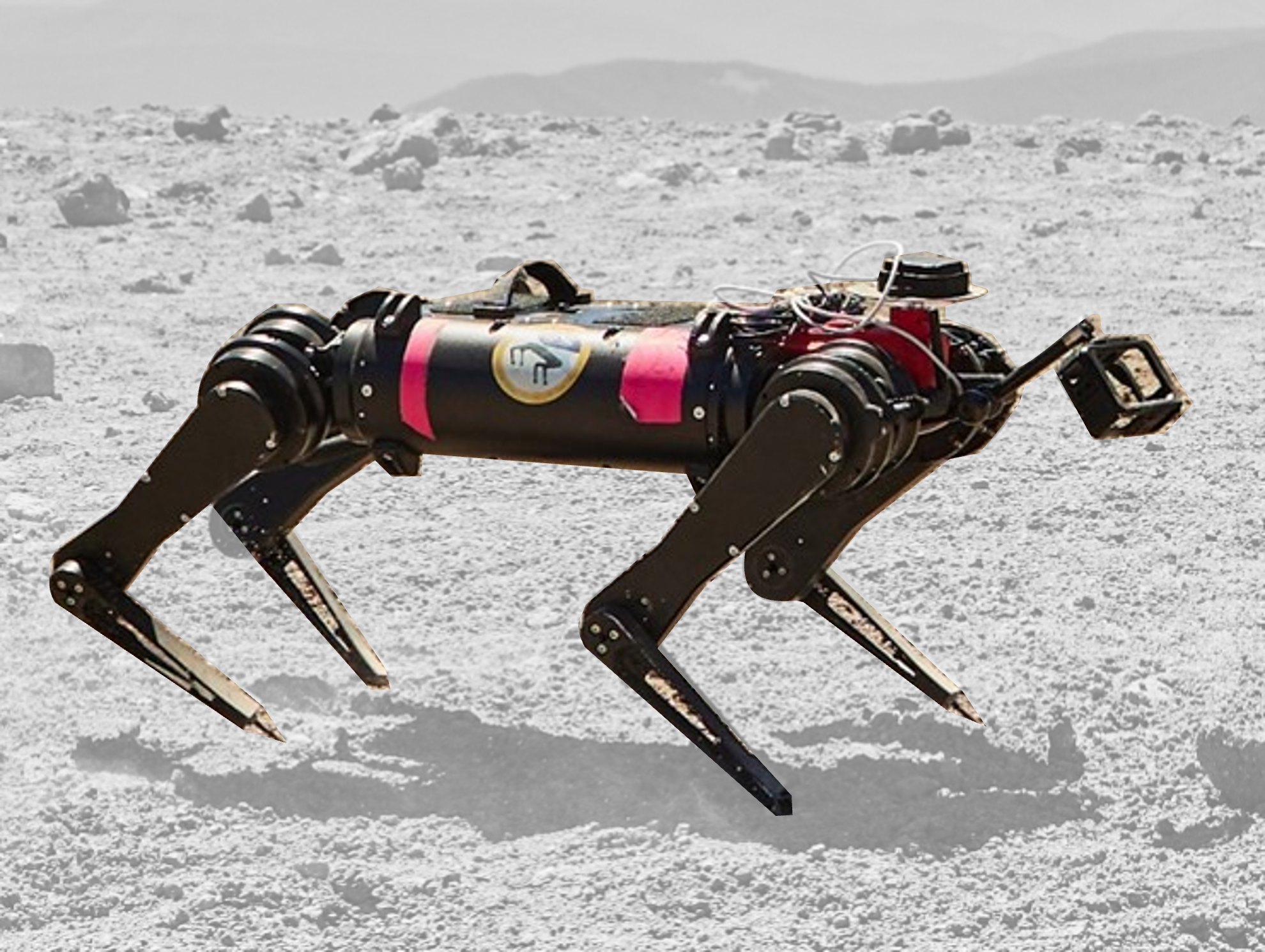}%
        {%
        \setlength{\fboxsep}{2pt}%
        \put(90, 3.5){\colorbox{white}{\textbf{\footnotesize{B}}}}
        }%
    \end{overpic}
        \phantomcaption
        \label{fig:intro_spirit}
    \\ \vspace{3pt} 
    \begin{overpic}[height=0.28\linewidth]{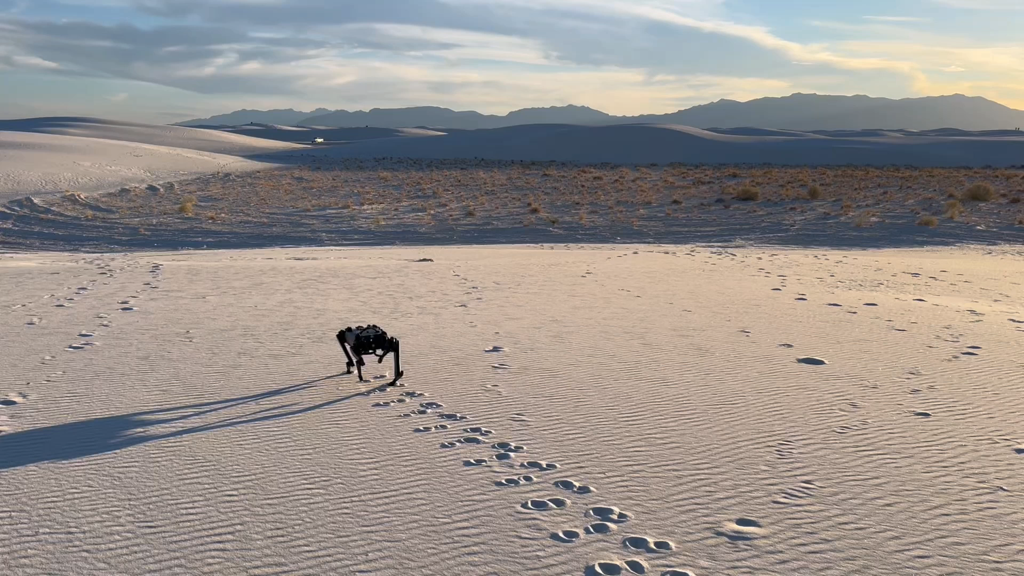}%
        {%
        \setlength{\fboxsep}{2pt}%
        \put(1.5, 3.5){\colorbox{white}{\textbf{\footnotesize{C}}}}
        }%
    \end{overpic}
        \phantomcaption
        \label{fig:intro_whitesands}
    \begin{overpic}[height=0.28\linewidth]{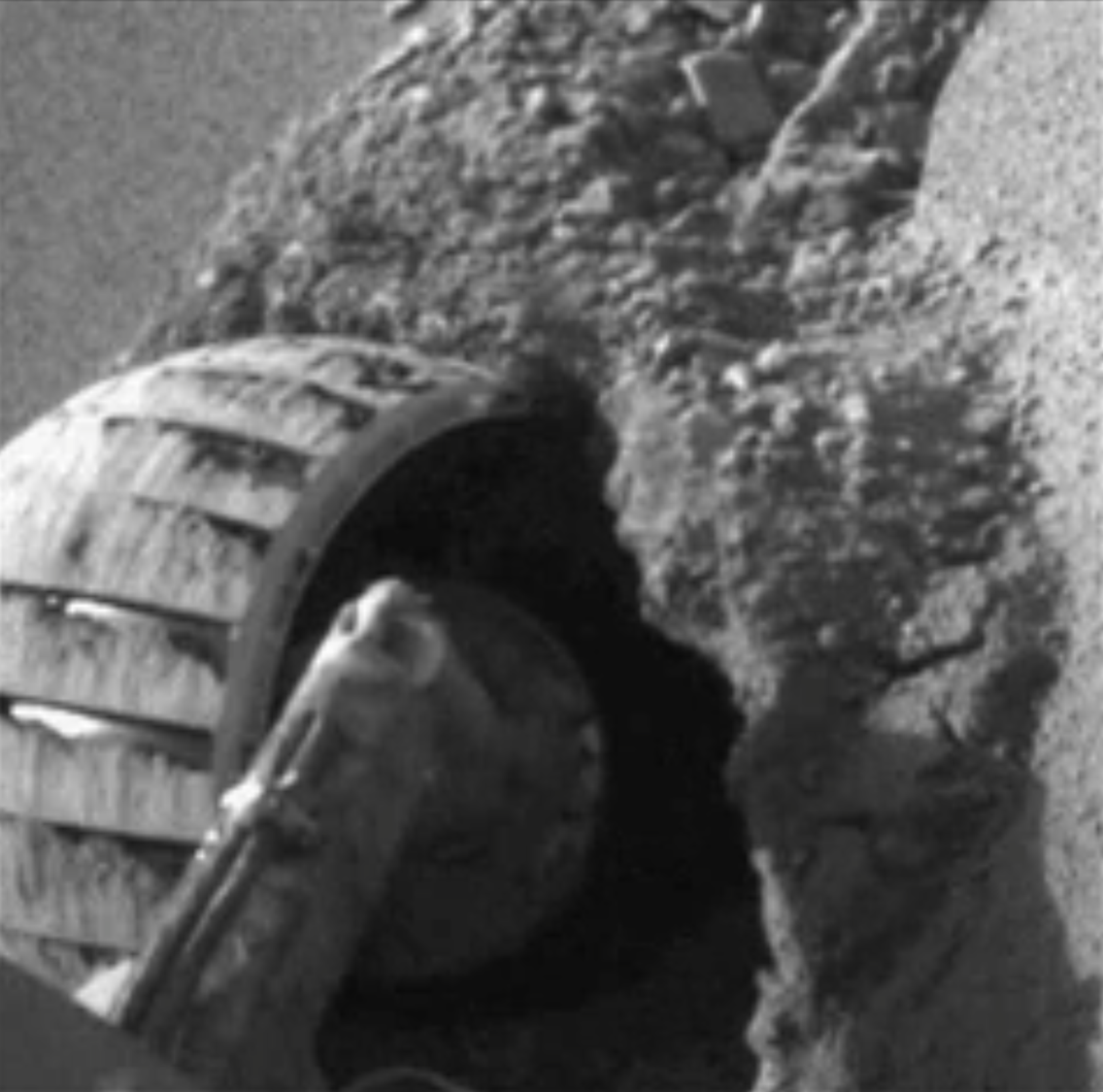}    %
        {%
        \setlength{\fboxsep}{2pt}%
        \put(82.5, 6.5){\colorbox{white}{\textbf{\footnotesize{D}}}}
        }%
    \end{overpic}
        \phantomcaption
        \label{fig:intro_wheel}
    \\ \vspace{3pt} 
    \begin{overpic}[width=0.8\linewidth]{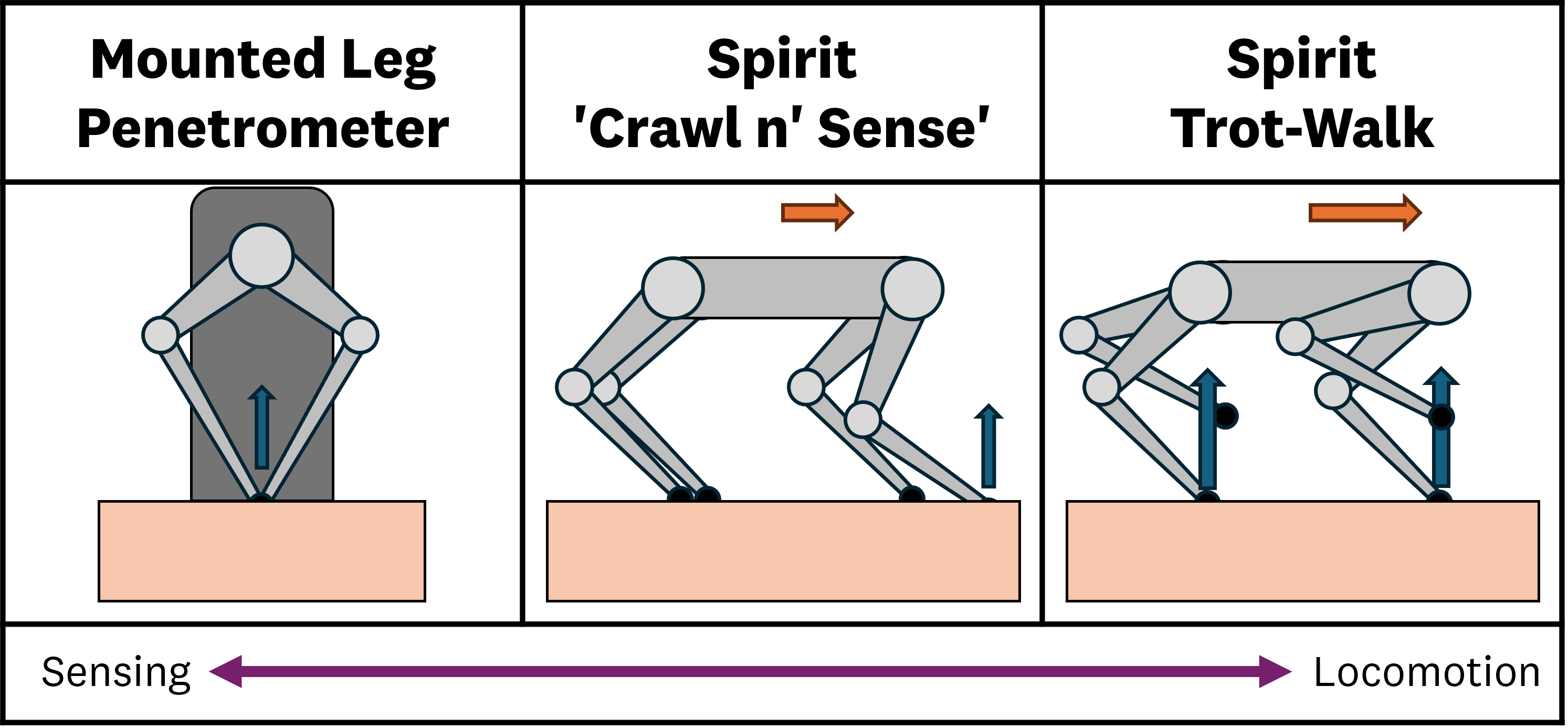}
    \put( 1.5, 30){{\textbf{\footnotesize{E.1}}}}
    \put(34.5, 30){{\textbf{\footnotesize{E.2}}}}
    \put(67.5, 30){{\textbf{\footnotesize{E.3}}}}
    \end{overpic}
    \phantomcaption
    \label{fig:intro_gaits}
\end{subfigure}
\caption{
Robotic legs act as effective sensing tools for terrain estimation and scouting.
(A) is a legged robot test stand using direct-drive, symmetric, 5-bar linkage leg \cite{bush2023robotic} as a penetrometer.
(B) is a quasi-direct-drive quadrupedal robot used for sensing while walking.
(C) shows the quadruped in a planetary analogue environment walking over and measuring the terrain.
(D) shows the sinkage in soft terrain of the NASA Opportunity rover's wheel\cite{nasajpl_rovers_nodate}.
(E.1-E.3) illustrates the hypothetical axis between sensing-oriented and locomotion-oriented gaits -- correlated with speed -- in particular showing the stationary mounted leg, the slow \crawlsense, and the \trotloco. 
}
\label{fig:intro}
\end{figure}

%% file: Sections/FieldResults.tex
\section{Robot-assisted planetary-analogue field exploration reveals complex regolith force responses}
\label{sec:fieldresults}
To investigate how force measurements can be used to characterize planetary substrates, we conducted field testing at Mt. Hood, Oregon, a lunar analogue environment (\cref{fig:fieldOver}).
At this site, a mounted robotic leg test stand (\cref{fig:intro_travleg,fig:intro_gaits}.1) was deployed along a geologist-defined transect to collect regolith force responses and characterize spatial variability of regolith mechanics. 
The six sampling locations (C.1-6) chosen by geologists along the transect, and their corresponding force responses, are shown in \cref{fig:fieldTrav:B} and \cref{fig:fieldTrav:C}.

Spanning a 10-meter segment, the data revealed rich variability in substrate responses. 
First, substrate strength varied significantly (rising from $5$ N/cm to $>$100 N/cm) over short distances.
At dry, loose regolith (C.1–C.2), the resistive force, $F_z$, increased slowly with penetration depth, $d$, indicating low strength (captured by the small slope of force vs. depth).
In contrast, wet or frozen regolith (C.3–C.5) produced much higher resistive force at shallower depth, reflecting substantially increased strength.
The ability to discern these strength variations would be crucial for assessing traversability and ensuring construction stability for planetary surface operations.

Second, substrate yield behavior also depend sensitively on regolith properties. 
In homogeneous regolith (C.1, C2, C3, C5), resistive force increased approximately linearly with penetration depth. 
However, in frozen regolith (C.6), rupture of the thin surface ice crust produced sudden force drops, distinct from the gradual yielding elsewhere.
Detecting such rupture signatures is crucial not only for safe mobility but also for identifying features of high planetary science interest, such as the presence of water ice or salty surface crusts. 

These spatial variations and abrupt changes in regolith mechanics would likely be overlooked with sparse sampling waypoints from a rover (often on the order of 10's or 100's of meters apart\cite{nasa_science_perseverance_nodate}) 
underscoring the importance of densely-distributed measurements.
These field observations motivate the use of a walking robot that can sense regolith mechanics at every step rather than relying on a few isolated probes. 
In addition, the measured regolith mechanics from our field mission also directly inform the subsequent laboratory experiments, where we examine how gait choice impacts a quadruped’s ability to capture both regolith strength and rupture behavior by walking.
%

\input{Figures/1-Motivation/motivation-v2.fig}

%% file: Figures/1-Motivation/motivation-v2.fig.tex
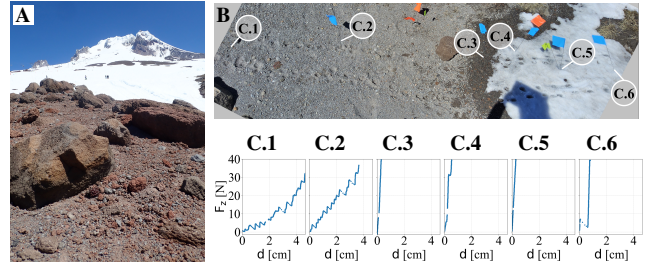
\begin{figure}[t]
\setlength{\abovecaptionskip}{5pt}
\centering
\vspace{6pt}
\begin{subfigure}[t]{\linewidth}
\begin{minipage}[b]{0.30\linewidth}   
    \begin{overpic}[angle=90,width=\linewidth]{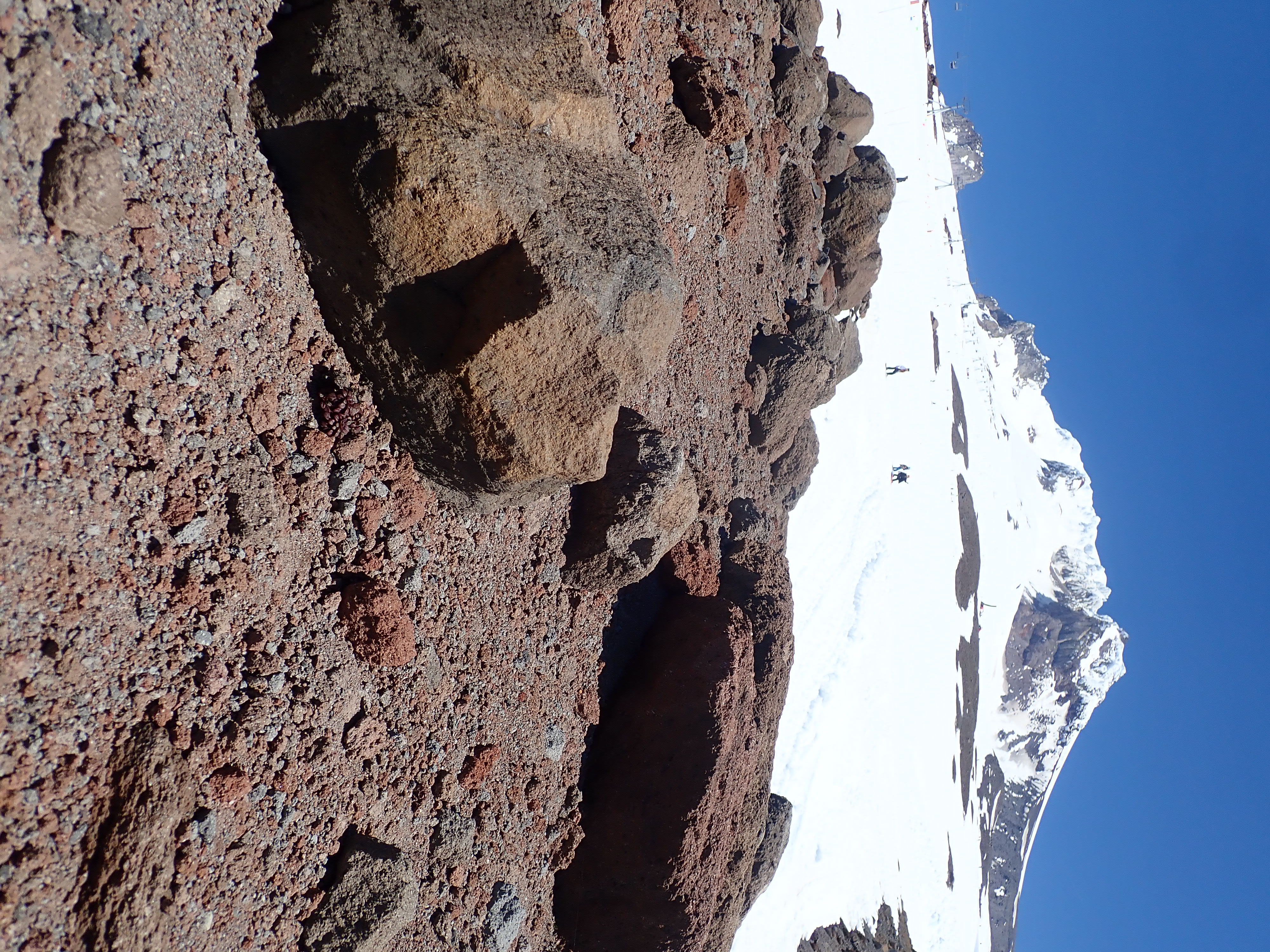}\put(-2.5, 92){
    {%
    \setlength{\fboxsep}{1pt}%
    \colorbox{white}{\textbf{\footnotesize A}}%
    }
} 
    \end{overpic}
    \phantomcaption
    \label{fig:fieldOver}
\end{minipage}
\begin{minipage}[b]{0.66\linewidth}
    \centering
    \begin{overpic}[width=\linewidth]{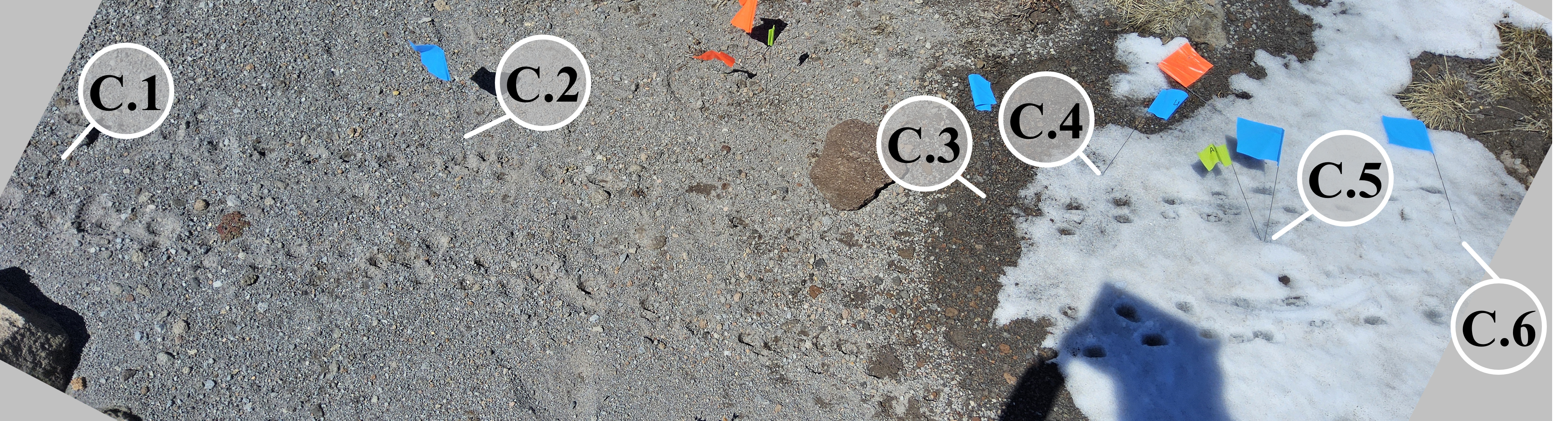}\put( 0.5, 22.5){\textbf{\footnotesize  B}} 
    \end{overpic}
    \phantomcaption
    \label{fig:fieldTrav:B}
    \\
    \vspace{3pt}
    \begin{overpic}[width=\linewidth]{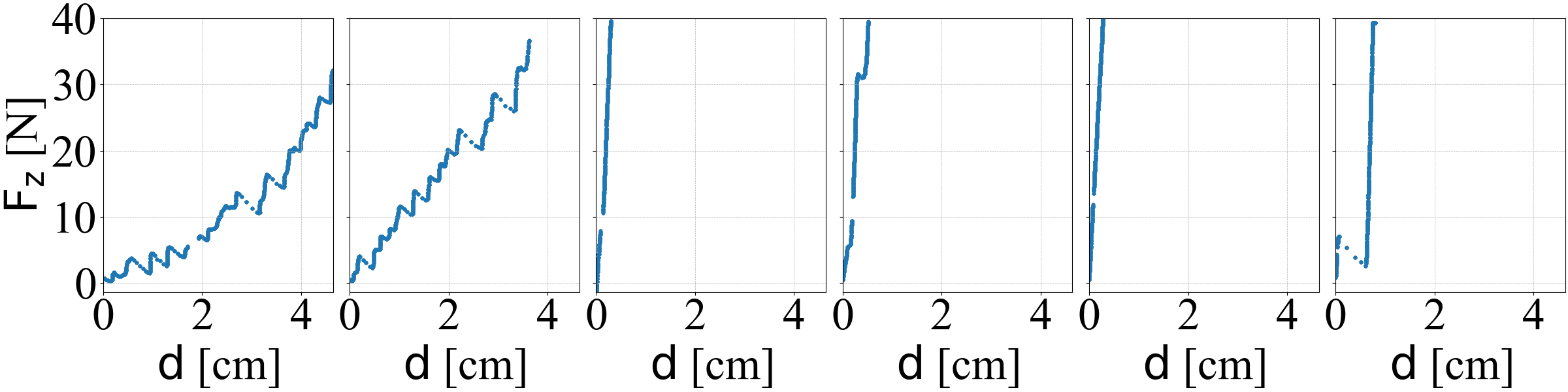}
        \put( 7.5, 26){\textbf{\footnotesize C.1}} 
        \put(23.35, 26){\textbf{\footnotesize C.2}} 
        \put(39.2, 26){\textbf{\footnotesize C.3}} 
        \put(55.05, 26){\textbf{\footnotesize C.4}} 
        \put(70.9, 26){\textbf{\footnotesize C.5}} 
        \put(86.75, 26){\textbf{\footnotesize C.6}} 
    \end{overpic}
    \phantomcaption
    \label{fig:fieldTrav:C}
\end{minipage}
\end{subfigure}
\caption{
Robot-leg assisted \insitu penetration force profiles across a snowy boundary in a planetary analogue environment, (A) (Mt. Hood, Oregon). (B) is an image of a nominal transect, and (C.1-6) show the force profiles at each of the labeled locations as measured with a robot-leg test stand. C.1-2 are dry, loose regolith, C.3 is wet regolith, C.4-6 are regolith with a thin layer of snow. There is a sharp increase in the strength of the material with the addition of moisture, and sharp brittle failures in the snow.}
\label{fig:MH}
\end{figure}

%% file: Sections/Methods.tex
\section{Methods to enable legged robots to measure terrain properties with every step}
\label{sec:methods}

In this section, we outline the methods used to investigate the sensing capabilities of a quadrupedal robot during locomotion. First, we introduce the robot used in this study and its unique features that enable the sensing capability (\cref{sec:methods-robot}).
Next, we describe the two gaits employed to investigate the effect of gait design on sensing performance: \crawlsense and  \trotloco  (\cref{fig:intro_gaits}.2-3), detailing their mechanisms, advantages, and limitations (\cref{sec:methods-gaits}). 
Then, we calculate an estimated ground plane for each step in each gait, which will be used to better estimate the terrain properties (\cref{sec:EstimatingDepth}). 
Finally, we present the experimental setup and data collection protocols, which involve testing the robot on two distinct substrates---homogeneous sand and heterogeneous synthetic crust---to validate the fidelity of the sensing data (\cref{sec:methods-setup}).

\subsection{Legged Robot}
\label{sec:methods-robot}
A quadrupedal robot (Ghost Robotics, Spirit 40) was used to investigate the gait effect. 
The robot weighs 12 kg, with a body length of 0.55 m. 
Each leg possesses 3 degrees of freedom (abduction, hip, and knee) (\cref{fig:intro_spirit}). For all four legs, the upper and lower linkages each measure 20 cm in length, with a cylindrical toe (4 cm diameter,  2 cm width).

All three leg actuators 
are quasi direct drive (\ie low gear ratio). 
The gearbox reductions are 6:1 on the abductor and hip actuators, and 12:1 on the knee.
This reduced gearbox ratio allows for actuator transparency \cite{katz2019mini,kau2019stanford}, a requirement for using measured motor torques for sensing, while still providing sufficient torque for effective locomotion.

The actuators are driven by a motor driver board (Ghost Robotics, Spirit 40 Motor Driver Triple Board). 
Gait-level control code is run on a microcontroller (Ghost Robotics, Spirit 40 mainboard) at 1 kHz. 
A computing module (Nvidia, TX2) sends high-level gait commands to the microcontroller via a custom MavLink protocol, logs low-level data, and communicates with the user through the Robot Operating System 2 (ROS2)  \cite{ros2} messaging protocol.



\subsection{Gaits}
\label{sec:methods-gaits}
To investigate the effect of gait design, especially locomotion speed, on sensing capabilities, we tested two representative gaits: a slower, sensing optimized gait which we refer to as \mbox{``\crawlsense''} (\cref{fig:intro_gaits}.2), and a faster, locomotion optimized gait which we refer to as \mbox{``\trotloco''} (\cref{fig:intro_gaits}.3). 
By comparing the sensitivity and accuracy of terrain sensing between the two representative gaits, we seek to understand how gait design impacts sensing performance while balancing locomotion speed and stability. 

\subsubsection{\crawlsense}
\label{sec:methods-crawl}

We developed the \crawlsense to optimize deformable terrain sensing accuracy. 
In \crawlsense, each step consists of two phases: a ``penetration'' phase, and a ``transition'' phase.
During the penetration phase, the robot maintains a stable tripod stance on three legs while the fourth leg acts as a ``penetrometer'' \cite{seweryn2014determining}, probing vertically into the deformable substrate with a constant speed. Meanwhile, motor position and current are recorded to compute ground reaction forces. 
 During the transition phase, the robot shifts its CoM forward and placing it above the support triangle of the next set of stance legs, following the heuristic described in \cite{focchiHighslopeTerrainLocomotion2017}.
Once the CoM reaches the desired placement, the next penetration leg begins its cycle: first recirculating and positioning its toe above the targeted penetration location,  then penetrating downward to initiate the next sensing phase.

For all \crawlsense trials tested in this study, a stride-wise gait frequency of 0.05 Hz and a step length of 10.5 cm were selected to achieve high-resolution sensing.
All other gait parameters, such as leg recirculation trajectory, max step length, and body height, were kept constant.
In addition, the penetration speed was kept low and constant (8.0 cm/s) to maximize data resolution during the penetration phase while minimizing granular inertial effects\cite{goldman2008scaling, gollub2011dynamical, qian2013walking}. 

A key feature of this gait design is the intentional separation of the penetration, recirculation, and body transition phases. This separation keeps the robot body still during the single-leg penetration, while the three supporting legs counteract reaction forces from the penetrating toe, thereby providing additional stability and reducing sensing noises.



\subsubsection{\trotloco}
\label{sec:methods-trot}
The \trotloco is a standard, locomotion-oriented, quadrupedal gait from the Ghost Robotics SDK \cite{uspto2020}.
In this gait, diagonal legs move synchronously, and the two diagonal pairs alternate between stance and swing phases. 
This trotting pattern is commonly observed in quadrupedal animals \cite{dagg1973gaits}, and serves as a benchmark gait in many state-of-the-art quadrupedal robots \cite{hutter2016anymal, katz2019mini}. 
Comparing this locomotion-oriented gait with the sensing-oriented \crawlsense gait offers key insights into the trade-offs between efficient locomotion and accurate terrain sensing.  

The \trotloco prioritizes locomotion efficiency and reactivity rather than precise force measurement. Unlike the \crawlsense, it does not separate sensing and locomotion phases. Instead, proprioception is used mainly to detect ground contact, without explicitly resolving penetration depth in deformable terrain. Because only two legs contact the ground at any given time, the gait introduces  additional challenges to obtain an accurate estimation of the ground plane, even on rigid surfaces.  
%

In our experiments, the stride frequency of \trotloco  was manually controlled to approximately 2 Hz via a joystick, with an actual range of $[1, 4]$ Hz. The higher cadence led to significantly shorter penetration durations (10-80 ms, compared to  1250 ms for \crawlsense), resulting in fewer usable penetration force measurements for estimating substrate strength. 
In addition, the higher leg touch down speeds in \trotloco can generate   inertial effects in granular media\cite{umbanhowar2010granular,qian2013walking}, adding nonlinear, impact-speed dependent, ground reaction forces.  
We discuss in \cref{sec:stiffness-detection} and \cref{sec:crust-detection} how these gait characteristics impact the terrain sensing performance. 





\subsection{Ground plane estimation for deformable substrate}
\label{sec:EstimatingDepth}
    Generally in legged robots, the contact between the toes and the ground is assumed to be rigid and non-yielding. However, when interacting with deformable terrains, accurately characterizing terrain strength requires estimating the penetration depth, defined as the perpendicular distance measured from the substrate surface to the submerged toe.
    Below we describe our method for estimating the ground surface, its normal vector, and the initial contact point for both gaits, to enable an estimation of penetration depth:
    \subsubsection{\crawlsense} 
    In the \crawlsense, a plane representing the ground surface is estimated using the contacts of the three supporting toes, in particular as the plane fit to those three points, whose normal vector is $\normalGround$ such that
    \begin{gather}
        v_{ij}\coloneq c_j - c_i, \quad 
        v_{ik}\coloneq c_k - c_i, \quad
        \normalGround = \dfrac{v_{ij} \times v_{ik}}{||v_{ij} \times v_{ik}||}
    \end{gather}
    where $c_i \in \mathbb{R}^3$ are the supporting toe contact locations and the pairwise differences, $v_{ij}$, give vectors whose cross product is the normal, $\normalGround$.
    \subsubsection{\trotloco} 
    In the \trotloco, two legs recirculate during penetration, so the previous method does not apply.
    Instead, our ground plane estimation algorithm uses numerical integration of the dynamics to carry forward (in time) an estimated position of the previous step's toe locations.
    A method similar to \cite{wang2023GroundPlaneIMU}---a regression-based extension of the \crawlsense plane estimation using the current and past toe locations (at least 3)---is then used for ground plane calculations. 
    \subsubsection{Correcting the ground plane frame with contact}
    On deformable substrates, the ground plane estimation using the supporting toe level is often inaccurate, due to the sinkage of supporting legs. We correct for this offset to improve the accuracy of penetration depth estimation. 
    Specifically, using the ground plane normal, $\normalGround$, we construct a frame whose $xy$-plane is perpendicular to $\normalGround$ and whose origin is the penetrating toe's position at the moment of contact. This position is determined using a force threshold for \crawlsense and via manual post-processing for \trotloco. By shifting the ground plane using this offset, the frame origin can be corrected to align with the actual substrate surface. 

\subsection{Experiment setup and data collection protocols}
\label{sec:methods-setup}
\input{Figures/3-ExperimentalSetup/experimental_setup.fig}
To evaluate the robot's ability in characterizing substrate strength and rupture behavior, we conducted two sets of controlled laboratory experiments. In the first experiment, the robot traversed across sand prepared with different compaction levels (\cref{fig:lab_transect_homo}). In the second set of experiment, the robot walked across a small sand section configured with surface crust. Both experiments were performed with the \crawlsense and \trotloco, to assess how gait design influences sensing performance.



\subsubsection{Experiment 1: homogeneous sand with varying compaction}
The goal of this experiment was to evaluate each gait’s ability to resolve differences in substrate strength across varying compaction levels.
To do so, we prepared three terrain units (TU's) of homogeneous damp sand at different compaction levels to vary substrate strength (\cref{fig:lab_transect_homo}).
TU 1 was set to medium compaction, TU 2 to low compaction, and TU 3 to high compaction (\cref{fig:lab_transect_homo}).
The medium condition was prepared by loosening and flattening the sand with a rake; the high-compaction condition was produced by tamping; and the low-compaction condition was generated by further fluffing with a rake\footnote{Due to the nature of fluffing and compacting sand, each TU's compaction slightly blends into its neighbors, leading to some ramping between them as reflected in the data(\cref{fig:bulk_data-crawldata}). }. 
Eight trials were performed for this set of experiments, including four trials with \crawlsense and four trials with \trotloco.

\subsubsection{Experiment 2: detecting ruptures in crusty terrain }
The goal of the second experiment was to evaluate the robot's ability to resolve abrupt changes in surface texture. 
In this experiment, TU 1, 2, 4, and 5 were configured as rigid (wooden) surfaces, while TU 3 was filled with either loose sand (300-micron
beach sand) topped with a surface crust (plaster tile crust substitute made with 1:1.5 plaster of Paris
to water) (\cref{fig:lab_transect_hetero,fig:setup:loose_sand,fig:setup:plaster}).
We had the robot traverse the setup to assess its ability to detect within-step variations in substrate strength---heterogeneous surface vs. subsurface, e.g., crust on sand.
Seven total trials were performed for this set of experiments, including four trials with \crawlsense and three trials with \trotloco. 

For both sets of experiments, robot joint position, velocity, and torque were logged at 1 kHz. To capture the ground truth of the robot's body and toe positions relative to the substrate boundaries and surfaces, a motion capture system was set up with four tracking cameras (OptiTrack Prime 13W) and two color video cameras (OptiTrack Prime Color), all running at 120 fps.
An additional video camera (GoPro HERO 12, recording at 60 fps) was used to capture wider side-view footage.
By integrating the onboard body pose estimation from the IMU and the kinematics, the toes' Cartesian position, velocity, and forces were calculated.

%% file: Figures/3-ExperimentalSetup/experimental_setup.fig.tex

\begin{figure}[tb!]
\centering
\vspace{6pt}
\begin{subfigure}[t]{0.98\linewidth}
    \begin{subfigure}[t]{0.95\linewidth}
        \centering
        \begin{overpic}[width=\linewidth]{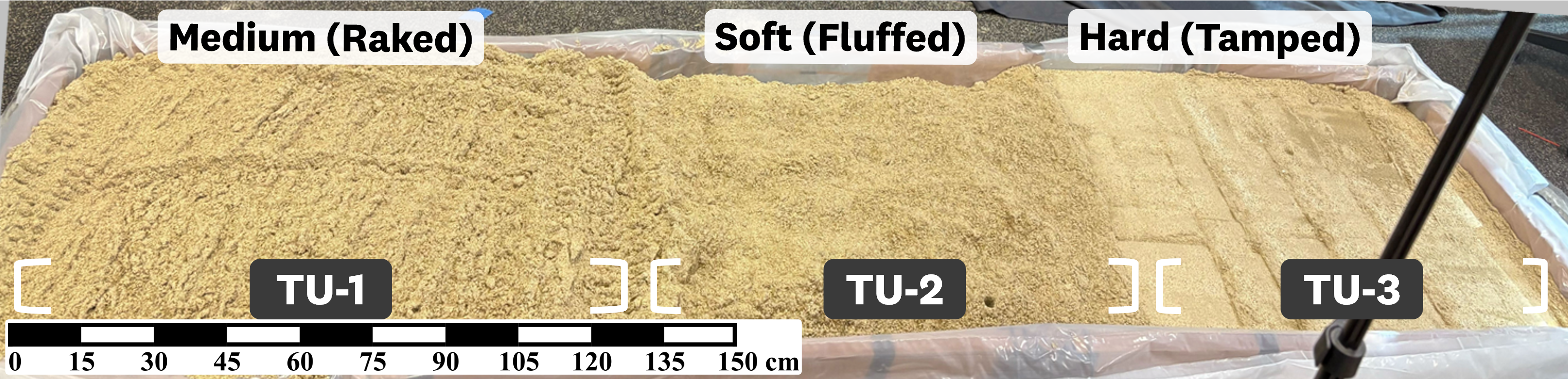}%
            {%
            \setlength{\fboxsep}{2pt}%
            \put(0.7, 20){\colorbox{white}{\textbf{\footnotesize{A}}}}
            }%
        \end{overpic}\\
        \vspace{5pt}
        \phantomcaption
        \label{fig:lab_transect_homo}
    \end{subfigure}
    \begin{subfigure}[t]{0.95\linewidth}
        \centering
        \begin{overpic}[width=\linewidth]{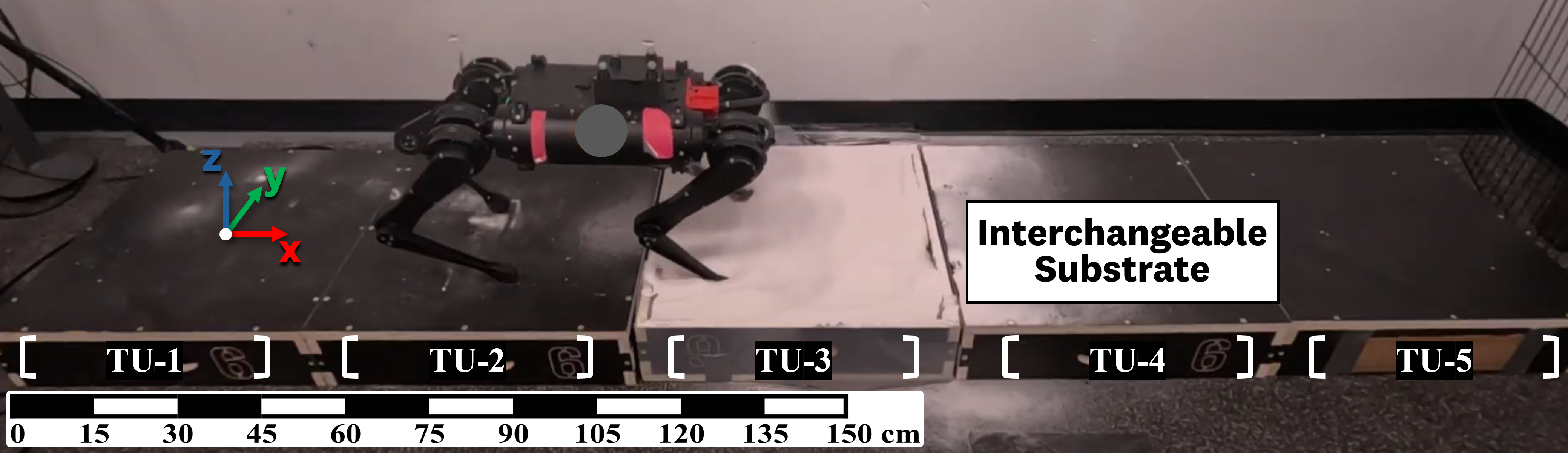}%
            {%
            \setlength{\fboxsep}{2pt}%
            \put(0.7, 25){\colorbox{white}{\textbf{\footnotesize{B}}}}
            }%
        \end{overpic}\\
        \vspace{3pt}
        \phantomcaption
        \label{fig:lab_transect_hetero}
    \end{subfigure}
    \begin{subfigure}[b]{0.99\linewidth}
        \centering
        \begin{overpic}[height=2.3cm]{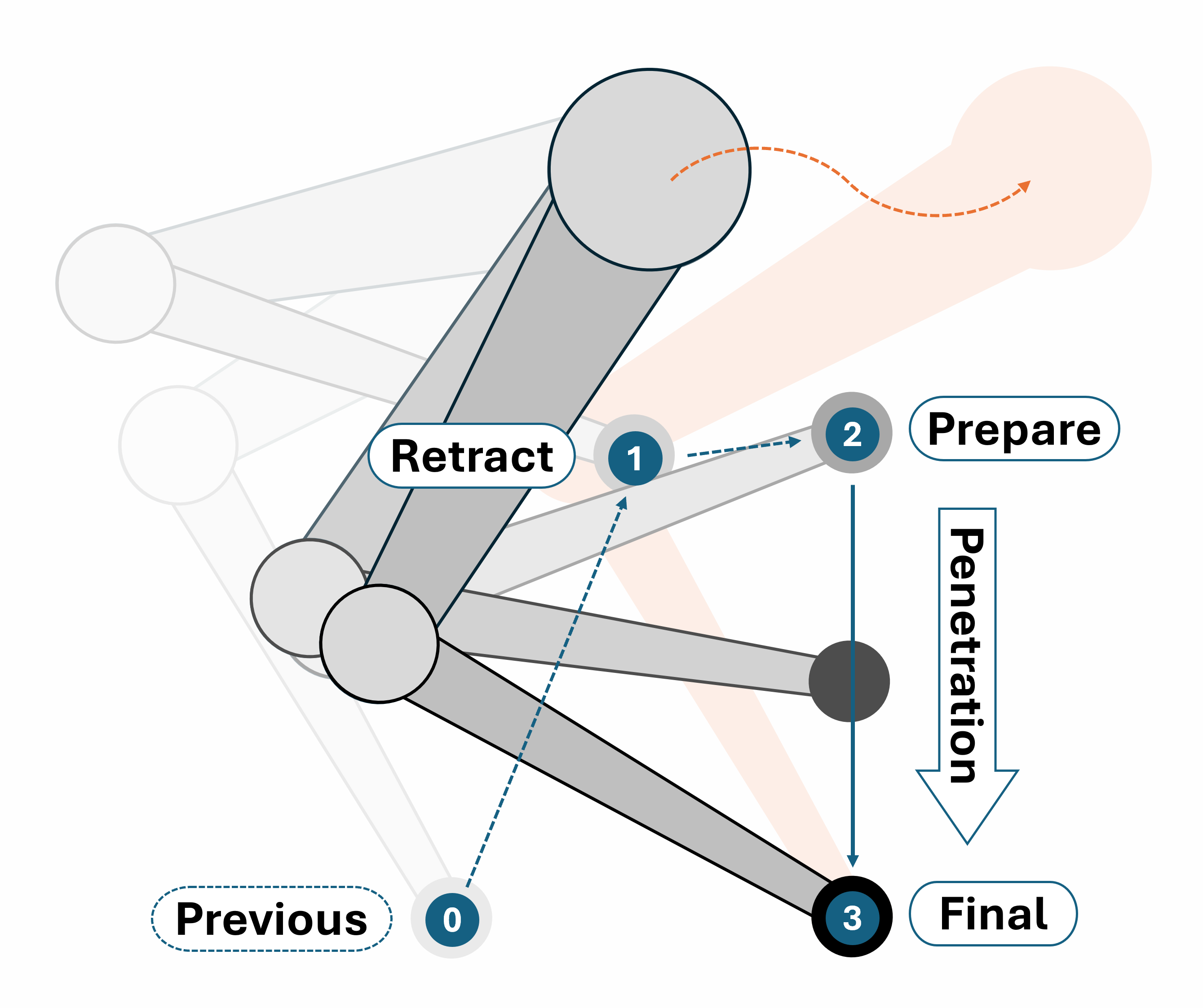}%
            {%
            \setlength{\fboxsep}{2pt}%
            \put(2, 75){\colorbox{white}{\textbf{\footnotesize{C}}}}
            }
        \end{overpic}
        \phantomcaption
        \label{fig:crawl_recirc}
        \begin{overpic}[height=2.3cm]{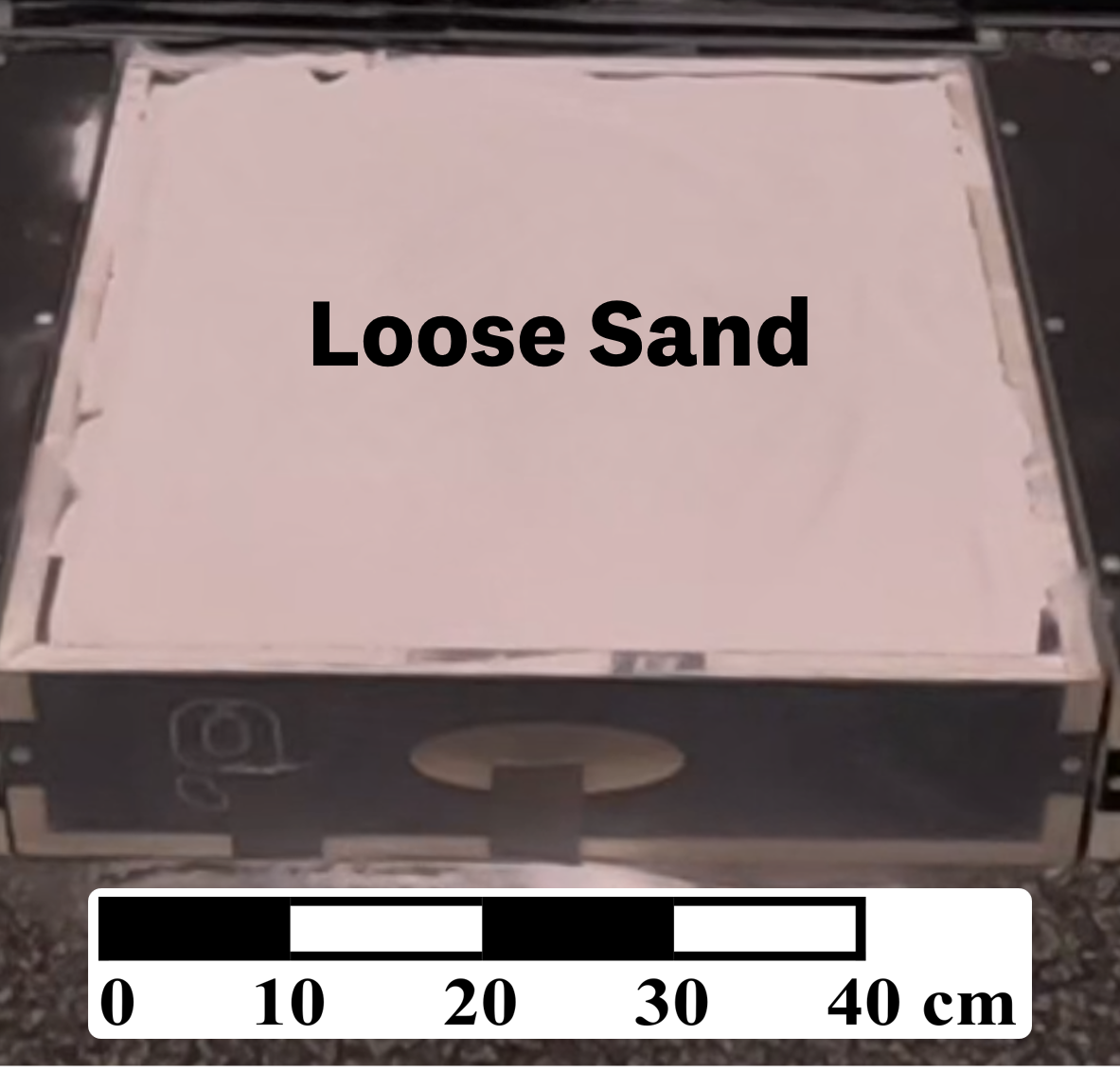}%
            {%
            \setlength{\fboxsep}{2pt}%
            \put(2, 83){\colorbox{white}{\textbf{\footnotesize{D}}}}
            }
        \end{overpic}
        \phantomcaption
        \label{fig:setup:loose_sand}
        \centering
        \begin{overpic}[height=2.3cm]{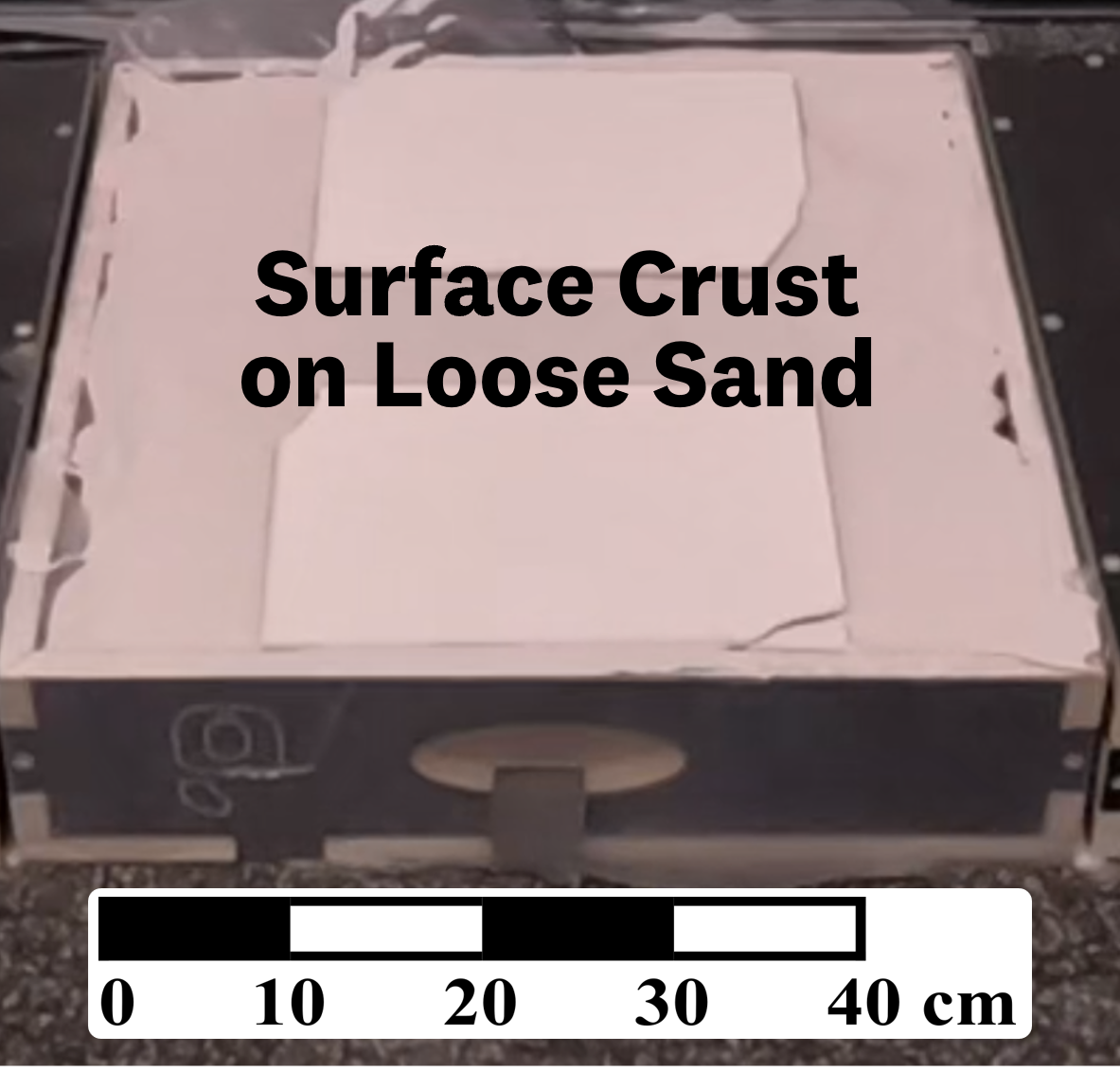}%
            {%
            \setlength{\fboxsep}{2pt}%
            \put(2, 83){\colorbox{white}{\textbf{\footnotesize{E}}}}
            }
        \end{overpic}
        \phantomcaption
        \label{fig:setup:plaster}
    \end{subfigure}
\end{subfigure}
\caption{%
Laboratory experimental setups.
The setup for Experiment 1 (A) is a 300cm $\times$ 80cm sandbox filled with construction sand.
The box was divided into three terrain units (TU) with different preparations: TU-1, raked (medium compaction); TU-2, fluffed (low compaction); and TU-3, tamped (high compaction).
The setup for Experiment 2 (B) is a 280cm $\times$ 56cm trackway made up of five wooden boxes/terrain units with the center one, TU-3, being interchangeable and filled with deformable substrates.
(C) describes the recirculation and penetration of the \crawlsense gait used to walk across.
All the terrain sensing happens during its penetration phase.
(D, E) Interchangeable deformable substrate in TU-3 of the second test track. 
(D) contains loosely packed sand.
(E) contains synthetic surface crusts on top of loosely packed sand. }

\label{fig:setup}
\end{figure}


%% file: Sections/Results.tex
\section{Effect of gait design on characterizing strength from homogeneous deformable substrates}\label{sec:stiffness-detection}

This section reports the examined effect of gait design on the accuracy of terrain strength characterization on homogeneous deformable substrates. We compare the sensing-oriented and locomotion-oriented gaits, and analyze the specific gait features that govern sensing performance.


\input{Figures/4-MethodDataAnalysis/method_analysis_new.fig}

\subsection{Determining penetration resistance from proprioception data during locomotion}\label{sec:homogeneous-sensing}
One of the key challenges in characterizing deformable terrain using signals from robot locomotion is relating the complex granular media responses observed during leg–terrain interaction to governing parameters such as compaction and strength.
Here, we focus on penetration resistance~\cite{nedderman1992statics,li2009sensitive}---a metric closely linked to compaction and substrate strength---and develop a method to estimate it from proprioceptively sensed leg forces.

\cref{fig:calculations:timeseries} shows the proprioceptive force measurements collected during a representative trial of \crawlsense from Experiment 1.
At the start of each shaded region, corresponding to the start of penetration (just before touchdown), the measured force rapidly increases from 0 to about 30 N, as the sand yield force increases with depth. 
Once the sand solidified~\cite{li2009sensitive,qian2015principles} to support the applied load, the force signal reached a plateau (e.g. $t=825.1$ and $t=848.0$). 
This marks the end of the penetration phase.
The gait then transitions into the support and recirculation phases (\cref{fig:crawl_recirc}), during which the measured forces vary as the other three legs manipulate body posture. 
At the end of the stride, the force returns to baseline before the next touchdown. 

The segment of the force signal most directly related to substrate strength is the interval between touchdown and the onset of sand solidification at 30 N, when the leg is penetrating vertically downward and the sand is continuously yielding.
\cref{fig:calculations:homo-time,fig:calculations:hetero-time} illustrate the proprioceptive force signal (blue) and the toe position (red) with this time segment highlighted. 
During this phase, the proprioceptively measured force increases linearly as penetration depth increases (\cref{fig:calculations:homo-penetration}). 
This dependence is consistent with prior granular physics studies~\cite{katsuragi2007unified, li2013terradynamics, kang2018archimedes}, which show that for slow intrusions into homogeneous granular media\cite{albert1999slow}, the resistance force primarily arises from the frictional pressure of the granular material, scaling proportionally with depth.
The slope of penetration resistive force \vs the penetration depth, known as the penetration resistance~\cite{stone2004local, stone2004getting}, can thus be used to quantify the strength of deformable granular substrates\cite{qian2015principles,qian2019rapid,ruck2023downslope}. 

To compute penetration resistance from proprioceptive sensing signals, we develop a method consisting of three key steps.
First, we determine the penetration depth, $d$, using the ground frame estimation method described in \cref{sec:EstimatingDepth}. 
Second, we identify the active penetration interval, $\timeROI~\coloneq~[t_j, t_{j+n}]$, for penetration resistance estimation. 
This interval corresponds to the linear increase in force during vertical penetration (shaded region in \cref{fig:calculations:homo-time} and cropped to in \cref{fig:calculations:homo-penetration}). 
For the \crawlsense gait, we applied fixed force thresholds of 15–30 N across all laboratory trials.
The lower bound was chosen conservatively to exclude frictional artifacts from slow leg recirculation, while the upper bound was set by the maximum penetration force allowed for stable stance.
In contrast, the dynamic contacts in \trotloco made these heuristics unreliable, so the penetration interval was manually identified during post-processing. 
Finally, within this interval, a linear regression was applied to estimate the penetration resistance, $\stiffness$, computed as the slope of the force-depth curve (\cref{fig:calculations:homo-penetration}), magenta dashed-line). 
\subsection{Effect of gait design on penetration resistance characterization}\label{sec:gait-effect-strength}
\cref{fig:bulk_data} shows the penetration resistance estimates from Experiment 1 for both \crawlsense (\cref{fig:bulk_data-crawldata}) and \trotloco (\cref{fig:bulk_data-trotdata}).
With both gaits, a clear difference in penetration resistance is observed between the low- and high-compaction sand. 
However, the full experiment sequence from medium to low to high compaction sand is more distinctly captured by the \crawlsense. 

Ground truth data collected with a mounted leg penetrometer along the center of the transect after the same terrain preparation measured $6.8 \pm 1.3$ N/cm in medium compaction sand, $3.4 \pm 1.3$ N/cm in low compaction, and $21.3 \pm 4.0$ N/cm in high compaction. %
The \crawlsense estimates closely tracked these ground-truth values in all three compaction zones.
Minor discrepancies between the two legs were likely due to differences in motor torque constant, joint friction, and kinematics at penetration between the LF and RF limbs.
Overall, the \crawlsense was able to reliably detect the shift between each compaction zone, both in individual leg estimates (\cref{fig:bulk_data}A, C, blue and red) as well as in aggregated fit using measurements from both legs  (\cref{fig:bulk_data}A, C, black).

In contrast, while \trotloco was able to qualitatively distinguish between the low and high compaction sand, the surface strength was consistently overestimated from all three compaction zones. In addition, the measurements from the same compaction region exhibited larger variance, particularly across the medium compaction sand. %
This is in part due to the inertial effects of the leg and the substrate.
Previous granular literature has found that substrate inertia contributes minimally to resistive forces during slow intrusions, such as in \crawlsense.
However, once the impact speed became large---such as in \trotloco where ground penetration speeds exceed 50 cm/sec---substrate inertia began to contribute significantly to granular resistive forces, resulting in a quadratic dependence of penetration force on impact speed~\cite{umbanhowar2010granular}.
This effect is further amplified by errors in ground force estimation at high recirculation speeds from the inertial effects of the leg itself.
The estimated forces in proprioceptive sensing contain both external (ground) forces and inertial forces, which can be difficult to disambiguate\cite{haddadin2017robotcollisionssurvey}.
The \trotloco implements a momentum observer to combat this inertial effect; however, our results suggested that momentum observers cannot fully compensate for the inertia effect across all leg linkages, and remain bandwidth-limited \cite{haddadin2017robotcollisionssurvey}.

These results suggested that both gait can qualitatively sense substrate strength differences. However, the \crawlsense provides significantly more reliable quantitative measurements of surface strength.

\section{Effect of gait design on detecting crusts in heterogeneous deformable substrates}\label{sec:crust-detection}
Unlike homogeneous sand, the proprioceptive force measurements from crusty sand (Experiment 2) exhibited sharp drops accompanied by sudden increases in penetration depth when the surface crust ruptures (\cref{fig:calculations:hetero-time,fig:calculations:hetero-penetration}).
This force drop occurs when the force exerted by the toe exceeds the binding force holding the crust together, causing the crust layer to fail abruptly\cite{langston2005experimental}.
Recent research has shown that the magnitude and slope of the force drop are indicative of the crust strength and brittleness~\cite{bush2024lpsc}, thus providing a pathway for legged robots to characterize heterogeneous substrates that include brittle surface layers.

In this section, we investigate how gait impacts the robot's capability of detecting texture and layering, such as surface crusts atop deformable sand. 

\subsection{Characterizing layering from proprioception data}\label{sec:heterogeneous-sensing}
To evaluate the robot's ability to detect ruptures in heterogeneous substrates, we first analyzed the proprioceptive force signals from each robot step, and developed a method for rupture detection. 


\input{Figures/5-DataFigure/bulk_data.fig}

To capture significant force drops, the raw proprioceptive force data (1kHz) was first filtered using a 4th-order Savitzy-Golay filter\cite{savitzkySmoothing1964} with a window of 60 ms.
This reduces the number of peaks and troughs in the data, while preserving the salient peaks and removing those due to noise or vibration.
We then identified troughs with significant prominence, by returning all cases where the force drop exceeded 5 Newtons.

Using this method, we extracted the steps where force-based rupture was detected, and compared them against ground truth substrate condition obtained from experimental video. From the video, each step was categorized into one of three conditions: ``rigid'', ``sand'', or ``crust'' (further divided into ``ruptured'' and ``intact''\footnote{During the heterogeneous substrate trials, some of the steps miss the tiles, instead entering the sand because we do not have $100\%$ coverage; these steps are considered to be on sand and not crust.}). 
Rupture detection accuracy was evaluated by matching the force-based flag with video ground truth as follows:
\begin{itemize}
    \item True Positive (TP): rupture detected when video shows “crust, ruptured.”
    \item True Negative (TN): no rupture detected when video shows “sand,” “rigid,” or “crust, intact.”
    \item False Positive (FP): rupture detected when video shows “sand,” “rigid,” or “crust, intact.”
    \item False Negative (FN): no rupture detected when video shows “crust, ruptured.”
\end{itemize}
These outcomes are summarized in \cref{fig:bulk_data-crawlcrust,fig:bulk_data-trotcrust}.


\subsection{Effect of gait design on detecting substrate rupture}\label{sec:gait-effect-rupture}
During the \crawlsense trials, the detections achieved a specificity (true negative rate) of $96\%$, with only 3 false positives out of 73 rupture-less steps. In addition, \crawlsense achieved a sensitivity (true positive rate) of $63\%$, missing 4 out of 11 rupture events. 

In contrast, the \trotloco trials showed a specificity of $12.5\%$, producing 56 false positives in 64 rupture-less steps, and a sensitivity of $100\%$, ``detecting" all the rupture events.
This sensitivity is misleading, however, since every step was flagged as a rupture, rendering unreliable detections.

These results suggested that the \crawlsense gait offers significant advantages over the \trotloco gait for detecting and resolving heterogeneity in substrate texture. This is likely due to the dedicated leg penetration phase in the \crawlsense. This dedicated phase provides a well-defined region for penetration resistance estimation, and allows for significantly higher data points to be gathered during penetration due to the slower penetration speed. The slow loading during the penetration phase also allows for better visibility of rapid force drops that occur during ruptures, which may be missed in \trotloco data. In contrast, the \trotloco's high penetration speeds, lack of a dedicated penetration phase, limits its ability to resolve rapid changes in penetration resistance forces, and introduced additional noises (\eg bounce) from high-speed impact. These effects significantly reduces the sensitivity and accuracy for \trotloco in characterizing substrate texture and layering. 

These observed differences in terrain sensing performance illustrated that robot gait characteristics had a large impact on terrain sensing performance, and underscore the inherent trade-off between locomotion and sensing performance. The \crawlsense prioritizes sensing, and enables high-resolution data and better accuracy of state estimation through quasi-static motion and carefully designed leg-surface interaction trajectory. In contrast, the \trotloco favors speed and coverage, allowing the robot to scout a larger area while collecting terrain information, but at the cost of fewer, lower-quality measurements and increased uncertainty in state estimation. 

%% file: Figures/4-MethodDataAnalysis/method_analysis_new.fig.tex
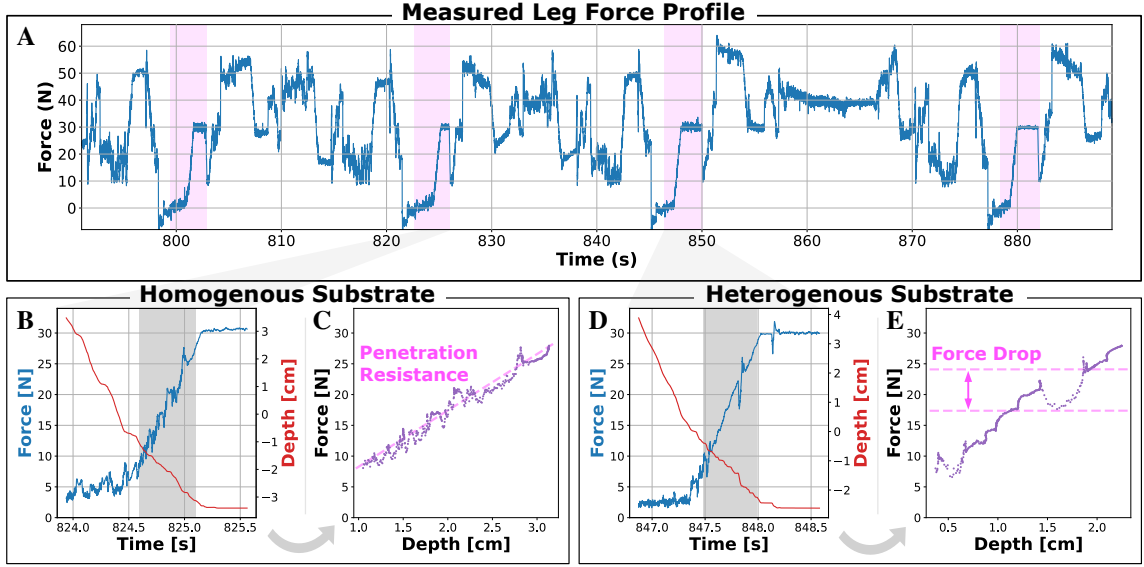
\begin{figure*}[t] 
\centering
\vspace{6pt}
\begin{overpic}[width=0.85\linewidth]{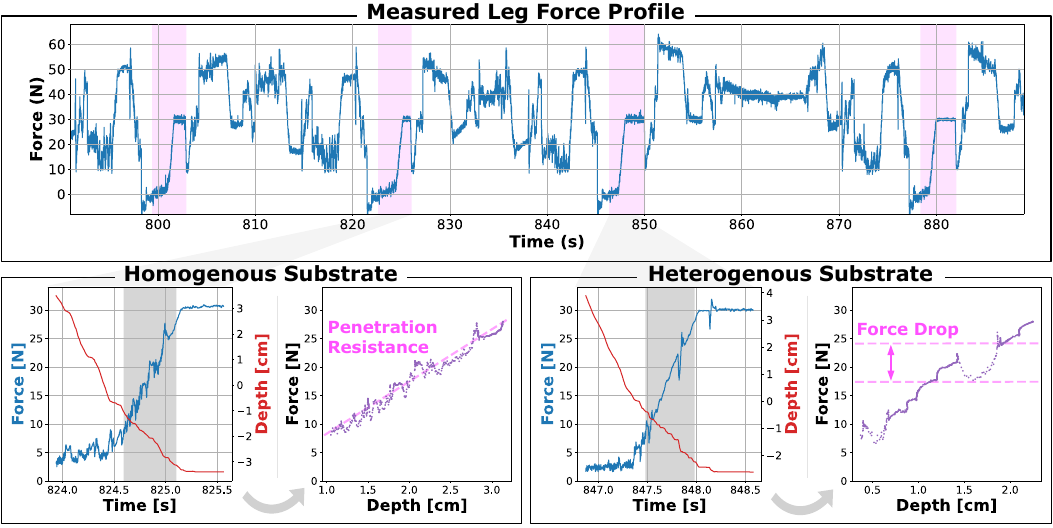}
            \put(1,46){\textbf{A}}
            \put(1,21){\textbf{B}}
            \put(27,21){\textbf{C}}
            \put(51.3,21){\textbf{D}}
            \put(77.5,21){\textbf{E}}
\end{overpic}
\caption{
Analyzing the robot proprioceptive force feedback for every step.
As the robot locomotes, it records the force profile at every toe; (A) shows the force \vs time profile for a \crawlsense trial in Experiment 2 across a crust-laden sand pit (specifically the right forelimb) with the penetration phase of each step highlighted.
(B) zooms onto one of those highlighted regions---a step into a homogeneous material--- and also includes estimated toe depth in red and a highlight region of interest (ROI) for the following computations.
Using the force and depth, (C) shows the output of a linear regression whose slope is the penetration resistance of the terrain as described in \cref{sec:homogeneous-sensing}. 
(D) is the same as (B) but for a footstep onto the simulated crust.
(E) shows the force drop associated with brittle failure and the measure we use to detect a rupture.
}
\label{fig:calculations}
\setcounter{subfigure}{0}
\refstepcounter{subfigure}\label{fig:calculations:timeseries}
\refstepcounter{subfigure}\label{fig:calculations:homo-time}
\refstepcounter{subfigure}\label{fig:calculations:homo-penetration}
\refstepcounter{subfigure}\label{fig:calculations:hetero-time}
\refstepcounter{subfigure}\label{fig:calculations:hetero-penetration}
\end{figure*}

%% file: Figures/5-DataFigure/bulk_data.fig.tex
\begin{figure*}[t!]
\centering
\vspace{6pt}
\begin{subfigure}[t]{\linewidth}
\centering
\setlength{\tabcolsep}{5pt} 
\setlength{\extrarowheight}{2pt}
\begin{tabularx}{0.93\linewidth}{@{}Y ? Z{0.44\linewidth} ? Z{0.28\linewidth}@{}}
    \textbf{Gait}
    &
    \textbf{Experiment 1}
    &
    \textbf{Experiment 2}
\\
\arrayrulecolor{gray!40} \hline \arrayrulecolor{black}
    \begin{subfigure}[t]{\linewidth}
        \includegraphics[width=\textwidth]{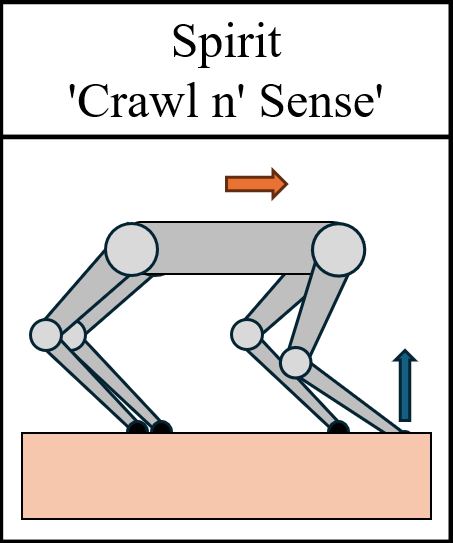}
    \end{subfigure}
&%
    \vspace{8pt}
    \begin{overpic}[width=\linewidth]{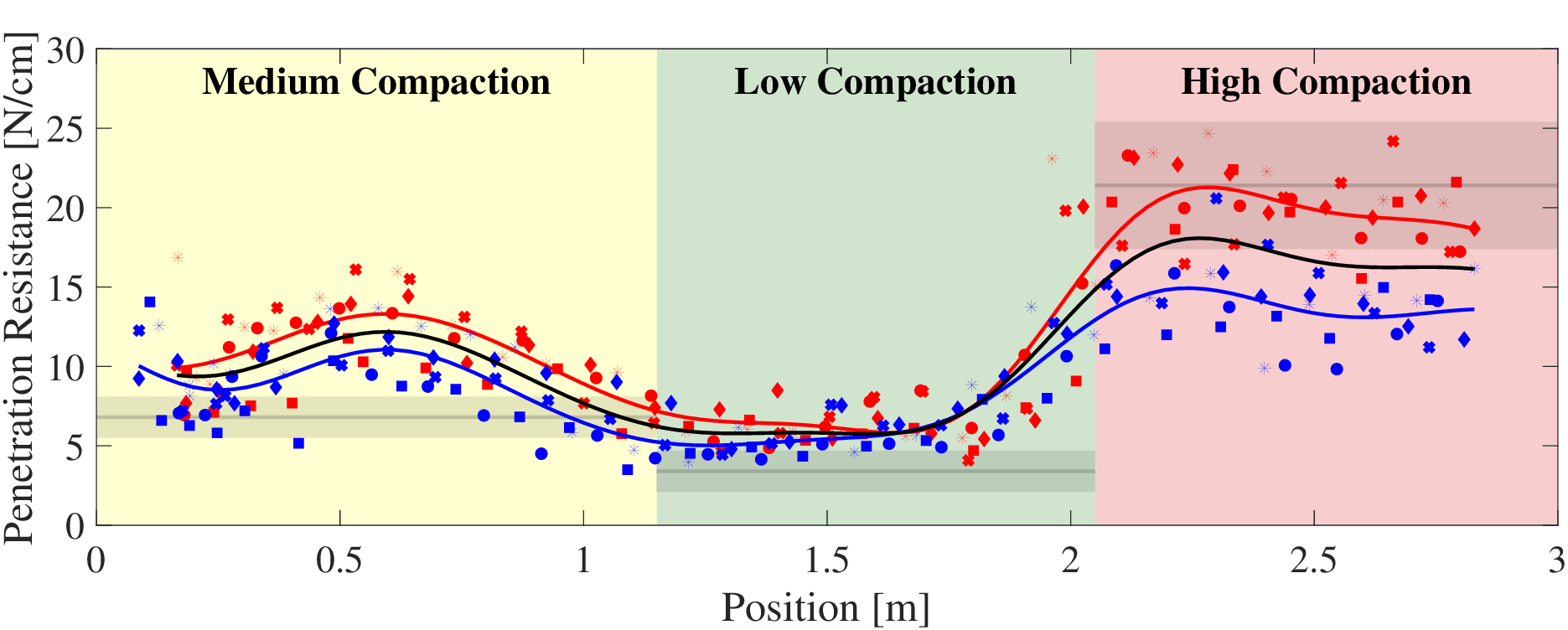}%
        {%
        \setlength{\fboxsep}{2pt}%
        \put(2, 40){\colorbox{white}{\textbf{\footnotesize{A}}}}
        }
    \end{overpic}
    \vspace{-4pt}
    \phantomcaption
    \label{fig:bulk_data-crawldata}
&%
    \vspace{8pt}
    \begin{overpic}[width=\linewidth]{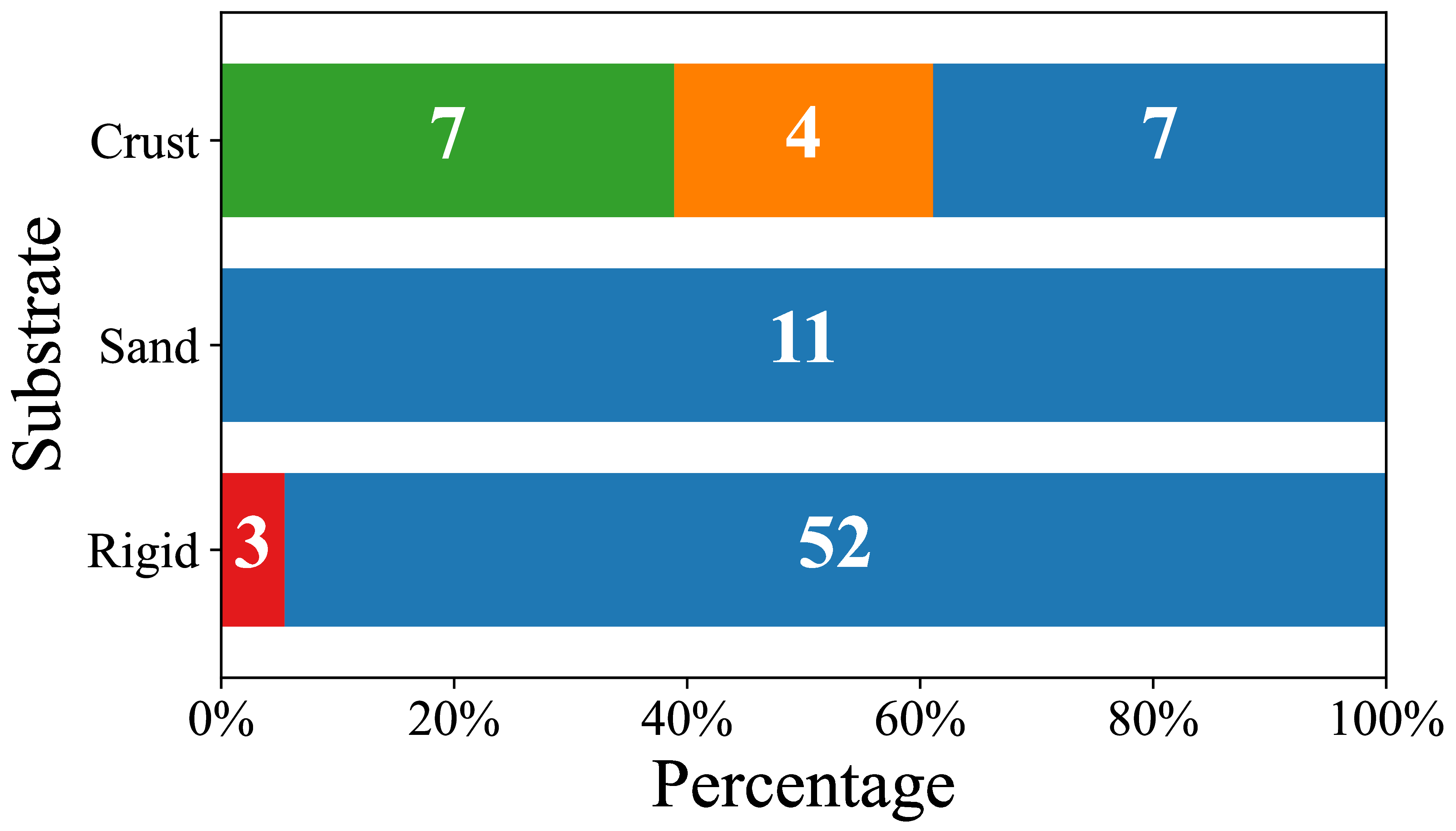}%
        {%
        \setlength{\fboxsep}{2pt}%
        \put(3,58){\colorbox{white}{\textbf{\footnotesize{B}}}}
        }
    \end{overpic}
    \vspace{-4pt}
    \phantomcaption
    \label{fig:bulk_data-crawlcrust}
\\
\arrayrulecolor{gray!40} \hline \arrayrulecolor{black}
    \begin{subfigure}[t]{\linewidth}
        \includegraphics[width=\linewidth]{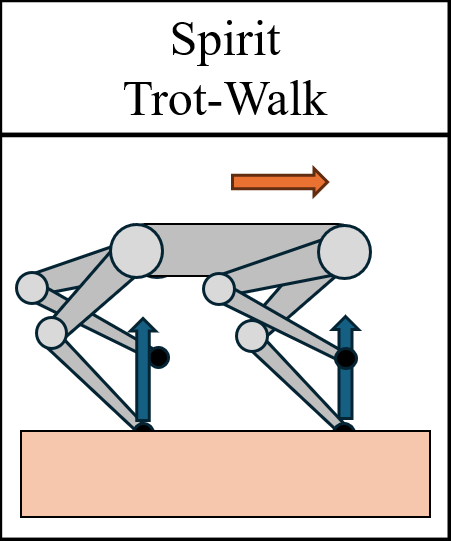}
    \end{subfigure}
&%
    \vspace{12pt}
    \begin{overpic}[width=\linewidth]{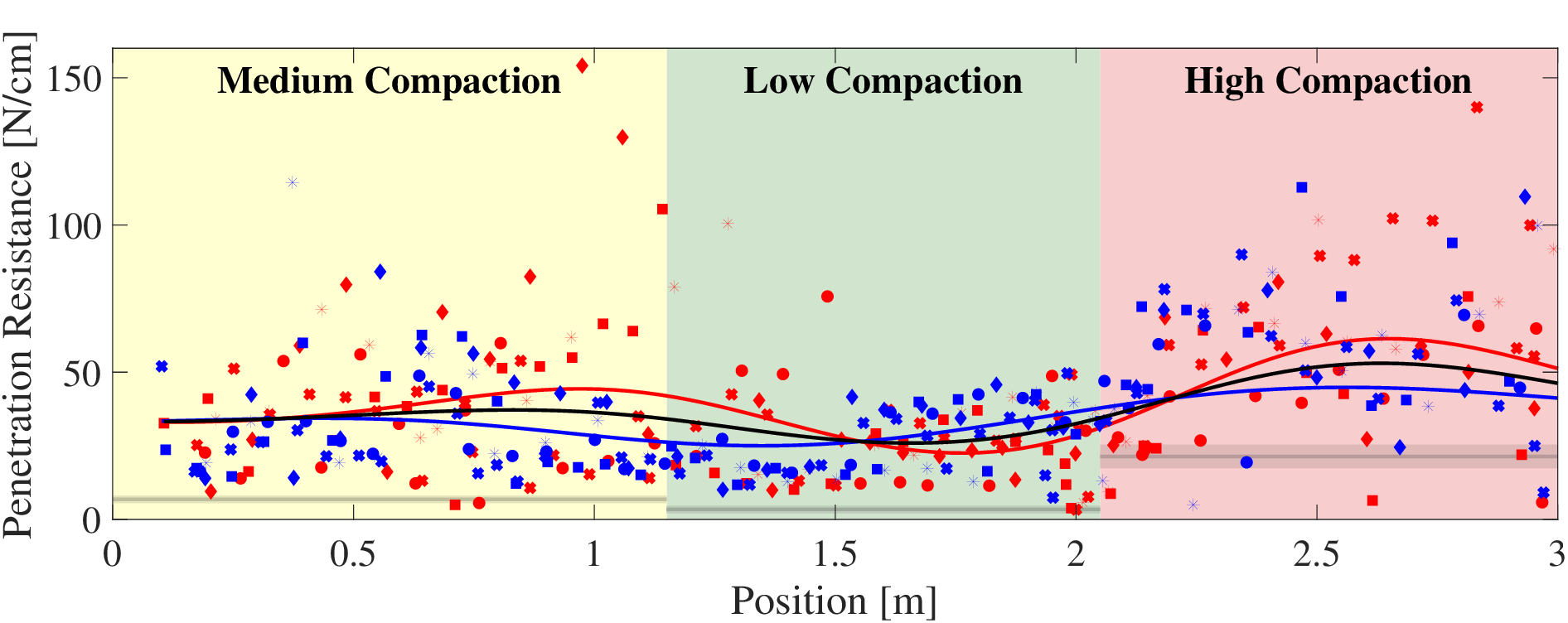}%
        {%
        \setlength{\fboxsep}{2pt}%
        \put(2, 40){\colorbox{white}{\textbf{\footnotesize{C}}}}
        }
    \end{overpic}
    \vspace{-10pt}
    \phantomcaption
    \label{fig:bulk_data-trotdata}
&%
    \vspace{12pt}
    \begin{overpic}[width=\linewidth]{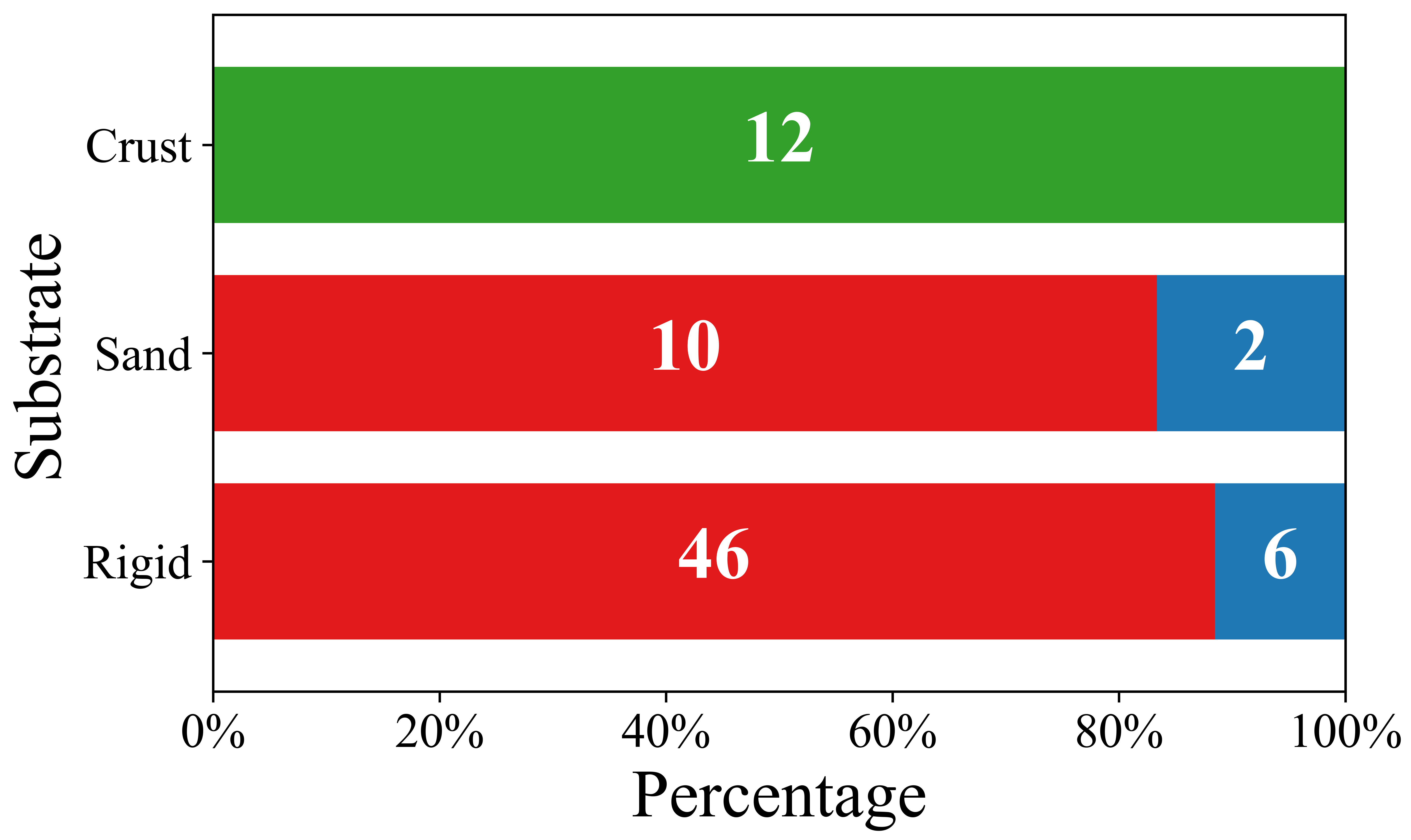}%
        {%
        \setlength{\fboxsep}{2pt}%
        \put(55,69){\makebox(0,0)[c]{\includegraphics[width=0.65\linewidth]{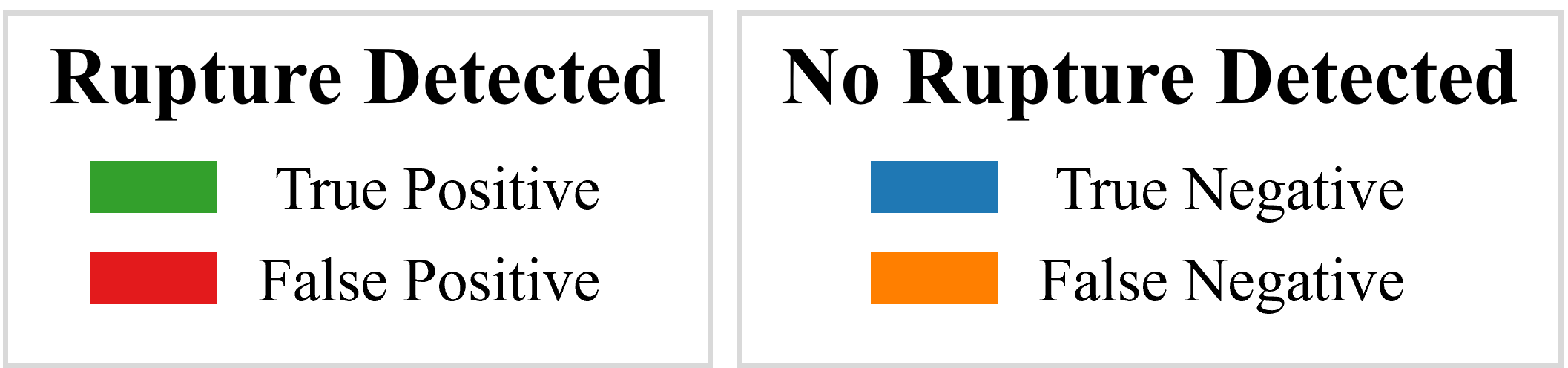}}}
        \put(3,60){\colorbox{white}{\textbf{\footnotesize{D}}}}
        }
    \end{overpic}
    \vspace{-10pt}
    \phantomcaption
    \label{fig:bulk_data-trotcrust}
\end{tabularx}
\end{subfigure}
\caption{
\textit{Results of lab trials.} 
Experiment 1 results are presented in  (A) and (C), the estimated penetration resistance is plotted against the position for each step during the \crawlsense (4 trials) and the \trotloco (4 trials), respectively. 
Color represents limb; blue for the right forelimb (RF) and red for the left forelimb (LF).
Marker shapes indicate different trials. 
The lines are solutions to Gaussian Process Regressions (GPR), aggregating the data for each limb (red and blue) and then averaging them (black); they reject noise/variation and provide interpolation between measured stiffness points.
(B) and (D) present Experiment 2 results, comparing the rupture detection results on plaster during \crawlsense (3 trials) and \trotloco (4 trials), expressing the number of detections for each case.
The robot flags any forelimb step with a force drop (\cref{fig:calculations:hetero-penetration}) of greater than 5~N.
Green, blue, red, and orange represent true positive (correctly detected a rupture), true negative (correctly did not detect a rupture), false positive (falsely detected a rupture), and false negative (falsely missed a rupture), respectively. 
}
\label{fig:bulk_data}
\end{figure*}

%% file: Sections/ConclusionDiscussions.tex
\section{Conclusion}
\label{sec:conclusiondiscussion}
In this study, we developed and evaluated methods that allow a quadrupedal robot to measure terramechanical properties of deformable substrates while locomoting. 
By designing and testing
a sensing-oriented custom gait (\crawlsense) against the standard locomotion-oriented gait (\trotloco), we demonstrated that gait design has a profound impact on accuracy and resolution of proprioceptive terrain sensing. 
Both gaits were able to detect transitions in substrate strength, however, the \crawlsense gait exhibited higher accuracy and repeatability, likely by incorporating a dedicated penetration phase and moving slowly
thus avoiding inertial and impact forces. 
Critically, the \crawlsense was able to resolve brittle surface crust ruptures, which the \trotloco failed to capture.

Our study provides a new method to enable proprioceptive characterization of terrain terramechanical properties during robot locomotion, which will allow us to build terrain maps quickly and efficiently while exploring and building on other worlds. 
These findings can also inform future gait design and selection for balancing sensing and locomotion performance, expanding robot-assisted scientific explorations to previously challenging-to-access environments. 



%% file: root.bbl
\begin{thebibliography}{10}
\providecommand{\url}[1]{#1}
\csname url@rmstyle\endcsname
\providecommand{\newblock}{\relax}
\providecommand{\bibinfo}[2]{#2}
\providecommand\BIBentrySTDinterwordspacing{\spaceskip=0pt\relax}
\providecommand\BIBentryALTinterwordstretchfactor{4}
\providecommand\BIBentryALTinterwordspacing{\spaceskip=\fontdimen2\font plus
\BIBentryALTinterwordstretchfactor\fontdimen3\font minus \fontdimen4\font\relax}
\providecommand\BIBforeignlanguage[2]{{%
\expandafter\ifx\csname l@#1\endcsname\relax
\typeout{** WARNING: IEEEtran.bst: No hyphenation pattern has been}%
\typeout{** loaded for the language `#1'. Using the pattern for}%
\typeout{** the default language instead.}%
\else
\language=\csname l@#1\endcsname
\fi
#2}}

\bibitem{https://doi.org/10.1029/2002JE002038}
\BIBentryALTinterwordspacing
J.~A. Crisp, M.~Adler, \emph{et~al.}, ``Mars exploration rover mission,'' \emph{Journal of Geophysical Research: Planets}, vol. 108, no. E12, 2003. [Online]. Available: \url{https://agupubs.onlinelibrary.wiley.com/doi/abs/10.1029/2002JE002038}
\BIBentrySTDinterwordspacing

\bibitem{doi:10.2514/6.2005-2525}
\BIBentryALTinterwordspacing
J.~Zakrajsek, D.~McKissock, \emph{et~al.}, \emph{Exploration Rover Concepts and Development Challenges}. [Online]. Available: \url{https://arc.aiaa.org/doi/abs/10.2514/6.2005-2525}
\BIBentrySTDinterwordspacing

\bibitem{CHHANIYARA2012115}
\BIBentryALTinterwordspacing
S.~Chhaniyara, C.~Brunskill, \emph{et~al.}, ``Terrain trafficability analysis and soil mechanical property identification for planetary rovers: A survey,'' \emph{Journal of Terramechanics}, vol.~49, no.~2, pp. 115--128, 2012. [Online]. Available: \url{https://www.sciencedirect.com/science/article/pii/S002248981200002X}
\BIBentrySTDinterwordspacing

\bibitem{bush2023robotic}
J.~Bush, Y.~Zhang, \emph{et~al.}, ``Robotic legs as novel planetary instrumentation to explore the mechanical properties of regolith,'' \emph{LPI Contributions}, vol. 2806, p. 1733, 2023.

\bibitem{nasajpl_rovers_nodate}
\BIBentryALTinterwordspacing
N.~JPL, ``\BIBforeignlanguage{en-US}{Rovers {Leaving} {Martian} {Sand} {Trap}}.'' [Online]. Available: \url{https://www.jpl.nasa.gov/videos/rovers-leaving-martian-sand-trap/}
\BIBentrySTDinterwordspacing

\bibitem{saranli2001rhex}
U.~Saranli, M.~Buehler, and D.~E. Koditschek, ``Rhex: A simple and highly mobile hexapod robot,'' \emph{The International Journal of Robotics Research}, vol.~20, no.~7, pp. 616--631, 2001.

\bibitem{kenneally2016design}
G.~D. Kenneally, A.~De, and D.~E. Koditschek, ``Design principles for a family of direct-drive legged robots.'' \emph{IEEE Robotics and Automation Letters}, vol.~1, no.~2, pp. 900--907, 2016.

\bibitem{katz2019mini}
B.~Katz, J.~Di~Carlo, and S.~Kim, ``Mini cheetah: A platform for pushing the limits of dynamic quadruped control,'' in \emph{2019 international conference on robotics and automation (ICRA)}.\hskip 1em plus 0.5em minus 0.4em\relax IEEE, 2019, pp. 6295--6301.

\bibitem{boston_dynamics_spot}
\BIBentryALTinterwordspacing
``Spot® - {The} {Agile} {Mobile} {Robot}.'' [Online]. Available: \url{https://www.bostondynamics.com/products/spot}
\BIBentrySTDinterwordspacing

\bibitem{arm2019SpaceBokDynamicLegged}
P.~Arm, R.~Zenkl, \emph{et~al.}, ``{{SpaceBok}}: {{A Dynamic Legged Robot}} for {{Space Exploration}},'' in \emph{2019 {{International Conference}} on {{Robotics}} and {{Automation}} ({{ICRA}})}.\hskip 1em plus 0.5em minus 0.4em\relax Montreal, QC, Canada: IEEE, May 2019, pp. 6288--6294.

\bibitem{ilhan2018autonomous}
B.~D. Ilhan, A.~M. Johnson, and D.~E. Koditschek, ``Autonomous legged hill ascent,'' \emph{Journal of Field Robotics}, vol.~35, no.~5, pp. 802--832, 2018.

\bibitem{qian2019rapid}
F.~Qian, D.~Lee, \emph{et~al.}, ``Rapid in situ characterization of soil erodibility with a field deployable robot,'' \emph{Journal of Geophysical Research: Earth Surface}, vol. 124, no.~5, pp. 1261--1280, 2019.

\bibitem{arm2023scientificexploration}
\BIBentryALTinterwordspacing
P.~Arm, G.~Waibel, \emph{et~al.}, ``Scientific exploration of challenging planetary analog environments with a team of legged robots,'' \emph{Science Robotics}, vol.~8, no.~80, p. eade9548, 2023. [Online]. Available: \url{https://www.science.org/doi/abs/10.1126/scirobotics.ade9548}
\BIBentrySTDinterwordspacing

\bibitem{hoeller2024anymalparkour}
\BIBentryALTinterwordspacing
D.~Hoeller, N.~Rudin, D.~Sako, and M.~Hutter, ``Anymal parkour: Learning agile navigation for quadrupedal robots,'' \emph{Science Robotics}, vol.~9, no.~88, p. eadi7566, 2024. [Online]. Available: \url{https://www.science.org/doi/abs/10.1126/scirobotics.adi7566}
\BIBentrySTDinterwordspacing

\bibitem{kenneally2018actuator}
G.~Kenneally, W.-H. Chen, and D.~Koditschek, ``Actuator transparency and the energetic cost of proprioception,'' \emph{International Symposium on Experimental Robotics}, 2018.

\bibitem{ruck2023downslope}
J.~G. Ruck, C.~G. Wilson, \emph{et~al.}, ``Downslope weakening of soil revealed by a rapid robotic rheometer,'' 2023.

\bibitem{qian2017ground}
F.~Qian, D.~Jerolmack, \emph{et~al.}, ``Ground robotic measurement of aeolian processes,'' \emph{Aeolian research}, vol.~27, pp. 1--11, 2017.

\bibitem{liu2025adaptive}
S.~Liu, J.~Tang, S.~Meng, and F.~Qian, ``Adaptive locomotion on mud through proprioceptive sensing of substrate properties,'' 2025.

\bibitem{jerolmack2006spatial}
D.~J. Jerolmack, D.~Mohrig, \emph{et~al.}, ``Spatial grain size sorting in eolian ripples and estimation of wind conditions on planetary surfaces: Application to meridiani planum, mars,'' \emph{Journal of Geophysical Research: Planets}, vol. 111, no. E12, 2006.

\bibitem{jerolmack2011sorting}
D.~J. Jerolmack, M.~D. Reitz, and R.~L. Martin, ``Sorting out abrasion in a gypsum dune field,'' \emph{Journal of Geophysical Research: Earth Surface}, vol. 116, no.~F2, 2011.

\bibitem{kostynick2022rheology}
R.~P. Kostynick, H.~Matinpour, \emph{et~al.}, ``Rheology of debris-flow materials is controlled by the distance from jamming,'' 2022.

\bibitem{Keenan2024CADRE}
J.-P. de~la Croix, F.~Rossi, \emph{et~al.}, ``Multi-agent autonomy for space exploration on the cadre lunar technology demonstration,'' pp. 1--14, 2024.

\bibitem{holladay2025LunarConstruction}
R.~Holladay, S.~Misra, \emph{et~al.}, ``Characterizing robot-ground interactions for autonomous lunar construction,'' 2025.

\bibitem{qian2013walking}
F.~Qian, T.~Zhang, \emph{et~al.}, ``Walking and running on yielding and fluidizing ground,'' \emph{Robotics: Science and Systems}, p. 345, 2013.

\bibitem{nasa_science_perseverance_nodate}
\BIBentryALTinterwordspacing
N.~Science, ``\BIBforeignlanguage{en-US}{Perseverance {Rover} {Location} {Map}},'' section: Mars 2020. [Online]. Available: \url{https://science.nasa.gov/mission/mars-2020-perseverance/location-map/}
\BIBentrySTDinterwordspacing

\bibitem{kau2019stanford}
N.~Kau, A.~Schultz, N.~Ferrante, and P.~Slade, ``Stanford doggo: An open-source, quasi-direct-drive quadruped,'' in \emph{2019 International conference on robotics and automation (ICRA)}.\hskip 1em plus 0.5em minus 0.4em\relax IEEE, 2019, pp. 6309--6315.

\bibitem{ros2}
O.~Robotics, ``Robot operating system 2 (ros 2),'' \url{https://www.ros.org}, 2023, accessed: 2025-01-22.

\bibitem{seweryn2014determining}
K.~Seweryn, K.~Skocki, \emph{et~al.}, ``Determining the geotechnical properties of planetary regolith using low velocity penetrometers,'' \emph{Planetary and Space Science}, vol.~99, pp. 70--83, 2014.

\bibitem{focchiHighslopeTerrainLocomotion2017}
\BIBentryALTinterwordspacing
M.~Focchi, A.~Del~Prete, \emph{et~al.}, ``\BIBforeignlanguage{en}{High-slope terrain locomotion for torque-controlled quadruped robots},'' \emph{\BIBforeignlanguage{en}{Autonomous Robots}}, vol.~41, no.~1, pp. 259--272, Jan. 2017. [Online]. Available: \url{http://link.springer.com/10.1007/s10514-016-9573-1}
\BIBentrySTDinterwordspacing

\bibitem{goldman2008scaling}
D.~I. Goldman and P.~Umbanhowar, ``Scaling and dynamics of sphere and disk impact into granular media,'' \emph{Physical Review E—Statistical, Nonlinear, and Soft Matter Physics}, vol.~77, no.~2, p. 021308, 2008.

\bibitem{gollub2011dynamical}
J.~P. Gollub, K.~N. Nordstrom, and D.~J. Durian, ``Dynamical heterogeneity in soft-particle suspensions under shear,'' 2011.

\bibitem{uspto2020}
\BIBentryALTinterwordspacing
G.~R. Corp, ``Ghost spirit 40 q-ugv (rev. 1.0),'' 2020. [Online]. Available: \url{www.ghostrobotics.io}
\BIBentrySTDinterwordspacing

\bibitem{dagg1973gaits}
A.~I. Dagg, ``Gaits in mammals,'' \emph{Mammal Rev}, vol.~3, no.~4, pp. 135--154, 1973.

\bibitem{hutter2016anymal}
M.~Hutter, C.~Gehring, \emph{et~al.}, ``Anymal-a highly mobile and dynamic quadrupedal robot,'' in \emph{2016 IEEE/RSJ international conference on intelligent robots and systems (IROS)}.\hskip 1em plus 0.5em minus 0.4em\relax IEEE, 2016, pp. 38--44.

\bibitem{umbanhowar2010granular}
P.~Umbanhowar and D.~I. Goldman, ``Granular impact and the critical packing state,'' \emph{Physical review E}, vol.~82, no.~1, p. 010301, 2010.

\bibitem{wang2023GroundPlaneIMU}
J.~Wang, Z.~Pan, \emph{et~al.}, ``Estimation of ground posture angle for quadruped robots based on imu,'' pp. 1269--1275, 2023.

\bibitem{nedderman1992statics}
R.~M. Nedderman \emph{et~al.}, \emph{Statics and kinematics of granular materials}.\hskip 1em plus 0.5em minus 0.4em\relax Cambridge University Press Cambridge, 1992, vol. 352.

\bibitem{li2009sensitive}
C.~Li, P.~B. Umbanhowar, \emph{et~al.}, ``Sensitive dependence of the motion of a legged robot on granular media,'' \emph{Proceedings of the National Academy of Sciences}, vol. 106, no.~9, pp. 3029--3034, 2009.

\bibitem{qian2015principles}
F.~Qian, T.~Zhang, \emph{et~al.}, ``Principles of appendage design in robots and animals determining terradynamic performance on flowable ground,'' \emph{Bioinspiration \& biomimetics}, vol.~10, no.~5, p. 056014, 2015.

\bibitem{katsuragi2007unified}
H.~Katsuragi and D.~J. Durian, ``Unified force law for granular impact cratering,'' \emph{Nature physics}, vol.~3, no.~6, pp. 420--423, 2007.

\bibitem{li2013terradynamics}
C.~Li, T.~Zhang, and D.~I. Goldman, ``A terradynamics of legged locomotion on granular media,'' \emph{science}, vol. 339, no. 6126, pp. 1408--1412, 2013.

\bibitem{kang2018archimedes}
W.~Kang, Y.~Feng, C.~Liu, and R.~Blumenfeld, ``Archimedes’ law explains penetration of solids into granular media,'' \emph{Nature communications}, vol.~9, no.~1, p. 1101, 2018.

\bibitem{albert1999slow}
R.~Albert, M.~Pfeifer, A.-L. Barab{\'a}si, and P.~Schiffer, ``Slow drag in a granular medium,'' \emph{Physical review letters}, vol.~82, no.~1, p. 205, 1999.

\bibitem{stone2004local}
M.~Stone, R.~Barry, \emph{et~al.}, ``Local jamming via penetration of a granular medium,'' \emph{Physical Review E}, vol.~70, no.~4, p. 041301, 2004.

\bibitem{stone2004getting}
M.~B. Stone, D.~P. Bernstein, \emph{et~al.}, ``Getting to the bottom of a granular medium,'' \emph{Nature}, vol. 427, no. 6974, pp. 503--504, 2004.

\bibitem{haddadin2017robotcollisionssurvey}
S.~Haddadin, A.~D. Luca, and A.~{Albu-Sch{\"a}ffer}, ``Robot {{Collisions}}: {{A Survey}} on {{Detection}}, {{Isolation}}, and {{Identification}},'' \emph{IEEE Transactions on Robotics}, vol.~33, no.~6, pp. 1292--1312, Dec. 2017.

\bibitem{langston2005experimental}
G.~Langston and C.~M. Neuman, ``An experimental study on the susceptibility of crusted surfaces to wind erosion: a comparison of the strength properties of biotic and salt crusts,'' \emph{Geomorphology}, vol.~72, no. 1-4, pp. 40--53, 2005.

\bibitem{bush2024lpsc}
J.~Bush, Y.~Zhang, \emph{et~al.}, ``Relating geotechnical properties of crusty regolith to morphology and mineralogy using a robotic leg rheometer,'' \emph{LPI Contributions}, 2024.

\bibitem{savitzkySmoothing1964}
\BIBentryALTinterwordspacing
A.~Savitzky and M.~J.~E. Golay, ``Smoothing and differentiation of data by simplified least squares procedures.'' \emph{Analytical Chemistry}, vol.~36, no.~8, pp. 1627--1639, 1964. [Online]. Available: \url{https://doi.org/10.1021/ac60214a047}
\BIBentrySTDinterwordspacing

\end{thebibliography}
